%% file: main.tex
\documentclass{article}

\usepackage{microtype}
\usepackage{graphicx}
\usepackage{subcaption}
\usepackage{booktabs}
\usepackage{multirow}

\usepackage{hyperref}

\usepackage[accepted]{icml2026}

\makeatletter
\renewcommand{\ICML@appearing}{\textit{Workshop on Graph Foundation Models
(GFM) at the $\mathit{43}^{rd}$ International Conference on Machine
Learning (ICML)}, Seoul, South Korea, 2026.}
\makeatother

\usepackage{amsmath}
\usepackage{amssymb}
\usepackage{mathtools}
\usepackage{amsthm}
\usepackage{algorithm}
\usepackage{algorithmic}
\usepackage{float}
\usepackage{placeins}
\usepackage{tikz}
\usetikzlibrary{positioning,fit,backgrounds,calc,arrows.meta,shapes}

\usepackage[capitalize,noabbrev]{cleveref}

\theoremstyle{plain}
\newtheorem{theorem}{Theorem}[section]
\newtheorem{proposition}[theorem]{Proposition}
\crefname{proposition}{Proposition}{Propositions}
\Crefname{proposition}{Proposition}{Propositions}

\theoremstyle{definition}

\theoremstyle{remark}

\newcommand{\Method}{\textsc{Grate}}

\newcommand{\Tau}{T}

\newcommand{\Gtrain}{G_{\mathit{train}}}
\newcommand{\Vtrain}{V_{\mathit{train}}}
\newcommand{\Rtrain}{R_{\mathit{train}}}
\newcommand{\Ttrain}{\Tau_{\mathit{train}}}
\newcommand{\Qtrain}{Q_{\mathit{train}}}

\newcommand{\Qobs}{Q_{\textsc{o}}}

\newcommand{\Qvalid}{Q_{\mathit{valid}}}

\newcommand{\Qtest}{Q_{\mathit{test}}}
\newcommand{\Ginf}{G_{\mathit{inf}}}
\newcommand{\Vinf}{V_{\mathit{inf}}}
\newcommand{\Rinf}{R_{\mathit{inf}}}
\newcommand{\Tinf}{\Tau_{\mathit{inf}}}
\newcommand{\Qinf}{Q_{\mathit{inf}}}

\newcommand{\topic}[1]{\par\noindent\textbf{#1}\space\ignorespaces}
\newcommand{\tightlistsetup}{%
  \setlength{\itemsep}{0.2ex}%
  \setlength{\parsep}{0pt}%
  \setlength{\topsep}{0.2ex}}
\newcommand{\subtabtitle}[1]{\par\textbf{#1}\par\vspace{0.2em}}
\newcommand{\subtabgap}{\par\vspace{0.8em}}

\icmltitlerunning{\Method{}: Temporal Extensions for Inductive KG Foundation Models via Gated Rotary Attention}

\makeatletter
\renewcommand\paragraph{\@startsection{paragraph}{4}{\z@}%
  {0.5ex \@plus 0.2ex \@minus .1ex}%
  {-1em}%
  {\normalfont\normalsize\bfseries}}
\makeatother

\begin{document}

\twocolumn[
  \icmltitle{\Method{}: Temporal Extensions for Inductive KG\\
    Foundation Models via Gated Rotary Attention}

  \icmlsetsymbol{equal}{*}

  \begin{icmlauthorlist}
    \icmlauthor{Jiaxin Pan}{stuttgart}
    \icmlauthor{Osama Mohammed}{stuttgart}
    \icmlauthor{Daniel Hern\'{a}ndez}{stuttgart}
    \icmlauthor{Steffen Staab}{stuttgart,southampton}
  \end{icmlauthorlist}

  \icmlaffiliation{stuttgart}{Institute for Artificial Intelligence, University of Stuttgart, Germany}
  \icmlaffiliation{southampton}{University of Southampton, UK}

  \icmlcorrespondingauthor{Jiaxin Pan}{jiaxin.pan@ipvs.uni-stuttgart.de}

  \icmlkeywords{temporal knowledge graphs, foundation models, rotary positional encoding, inductive learning, graph neural networks}

  \vskip 0.3in
]

\printAffiliationsAndNotice{}

\input{sections/abstract}
\input{sections/intro}
\input{sections/related}

\input{sections/preliminaries}

\input{sections/background}
\input{sections/method}
\input{sections/benchmark}
\input{sections/experiments}

\input{sections/conclusion}

\section*{Acknowledgements}

We gratefully acknowledge computing time on the HoreKa supercomputer at the National High-Performance Computing Center at KIT (NHR@KIT), jointly funded by the German Federal Ministry of Education and Research (BMBF), the Ministry of Science, Research, and the Arts of Baden-W\"{u}rttemberg, and the German Research Foundation (DFG). Jiaxin Pan and Osama Mohammed acknowledge funding from the EU Chips Joint Undertaking (GA 101140087, SMARTY) and from the BMBF sub-project 16MEE0444. Daniel Hern\'{a}ndez is funded by the German Research Foundation (DFG) -- SFB 1574 -- 471687386.

\input{sections/limitations}

\bibliographystyle{icml2026}
\bibliography{references}

\clearpage
\appendix

\input{sections/app_sweep_full}

\input{sections/app_implementation}
\input{sections/app_datasets}
\input{sections/app_complexity}
\input{sections/app_heldout_transductive}
\input{sections/app_transductive_full}
\input{sections/app_baseline_inapplicability}
\input{sections/app_theoretical}

\input{sections/app_beta_sensitivity}

\end{document}

%% file: sections/abstract.tex
\begin{abstract}
Knowledge graph (KG) foundation models such as
\textsc{Ultra} and \textsc{Trix} generalize to unseen entities and
relations. Extending this transferability to temporal knowledge
graphs (TKGs) remains challenging: most temporal models tie their
parameters to dataset-specific vocabularies or hyperparameter
selection.
We propose \Method{}, a parameter-free temporal message function
that extends static KG foundation models with temporal reasoning.
Given a temporal query, \Method{} encodes each message by the
relative time gap between the corresponding fact and the query
and uses a query-conditioned gate to weigh each message by its
temporal and semantic relevance to the query.
To evaluate the model's temporal transferability, we construct 16
inductive transfer benchmark variants covering different levels
of inductive difficulty.
Across these benchmarks and two additional held-out forecasting
datasets, a single jointly pretrained \Method{} checkpoint
improves over static KG foundation models and outperforms FITTER
on link prediction tasks without dataset-specific hyperparameter
tuning.
\end{abstract}

%% file: sections/intro.tex
\section{Introduction}
\label{sec:intro}

Knowledge graph (KG) foundation models predict missing facts in entirely new KGs whose entities and relations were not part of the training data. Recent models such as \textsc{Ultra} \citep{galkintowards} and \textsc{Trix} \citep{zhangtrix} learn vocabulary-agnostic structural representations that transfer across independently constructed KGs. However, these methods do not model temporal information of facts.

Many knowledge graphs record when facts hold by associating each fact with a timestamp, forming temporal knowledge graphs (TKGs). For example, a political event graph may record (\emph{Obama, negotiate with, Merkel, 2014-03-24}). Temporal reasoning over such graphs is important in many practical settings. Enterprise KGs from different organisations contain distinct entities and relations, yet procurement cycles, contract renewals, and project milestones can exhibit similar temporal patterns. Likewise, digital twins of different manufacturing sites describe different equipment and processes, while maintenance cycles and operational state transitions may follow similar dynamics. Learning reusable temporal patterns across TKGs could therefore enable temporal reasoning to transfer to new graphs without retraining for each target.

Existing TKG models \citep{tltcomplexzhang2022along, li2021temporal, sun2021timetraveler} learn representations tied to their training entities, relations, and timestamps, and must be retrained for each new graph. FITTER \citep{pan2026fittervocabularyagnosticcrossdomaininference} enables inductive reasoning across disjoint entity, relation, and timestamp vocabularies by encoding ordered timestamp indices with sinusoidal functions. However, its temporal encoding requires selecting a dataset-specific frequency hyperparameter. To address this, we aim to extend KG foundation models for temporal reasoning without target-specific temporal adaptation.

However, this introduces two challenges. First, \emph{timestamps should be represented in a transferable and dataset-independent way}. KG foundation models avoid entity- and relation-specific representations; introducing absolute timestamp embeddings or target-specific temporal parameters would reintroduce a dependency on dataset-specific parameters. Source and target TKGs may contain disjoint timestamps and operate over different temporal ranges, so temporal information must be encoded without timestamp-specific parameters.

Second, \emph{predicting a missing temporal fact requires considering both structural and temporal relevance}. In a static KG, neighboring facts coexist without temporal ordering and are aggregated according to their structural relevance to the query. In a TKG, supporting facts occur at different temporal distances from the query and should not contribute equally. Temporally proximate facts often provide immediate evidence, but proximity alone is insufficient: a temporally distant fact can remain informative when strongly related to the query, while a recent but unrelated fact may contribute little. Temporal reasoning therefore requires jointly considering \emph{when} a fact occurred and \emph{how relevant} it is to the query.

We address these challenges with \Method{} (Gated Rotary Attention for Temporal Encoding), a parameter-free temporal message function that integrates directly into NBFNet-style \citep{zhu2021neural} KG foundation models. To represent timestamps without dataset-specific parameters, \Method{} encodes each message by its relative time gap to the query using rotary position embeddings. To account for both temporal and semantic relevance, it applies a query-conditioned gate that weighs each message by its temporal and semantic alignment with the query. Both operations use only relative time differences and existing hidden representations, introducing no additional learnable parameters or target-specific temporal adaptation. \Method{} thereby extends KG foundation models with temporal reasoning while preserving their vocabulary-agnostic transferability.

To evaluate this setting systematically, we construct a suite of 16 inductive transfer benchmarks covering different levels of entity, relation, and timestamp overlap between training and inference. These range from partially disjoint settings that model the emergence of new vocabulary within an evolving TKG to fully disjoint settings that model deployment on an independently constructed graph. The benchmarks cover both interpolation, where queries fall within the observed temporal range, and extrapolation, where queries concern future timestamps. We additionally evaluate zero-shot transfer on two independently held-out datasets. Across these settings, a single jointly pretrained \Method{} checkpoint achieves stronger link prediction than both static KG foundation models and outperforms FITTER without target-specific hyperparameter selection. These results demonstrate that temporal reasoning can be learned jointly and transferred across independently constructed TKGs.

%% file: sections/related.tex
\section{Related Work}
\label{sec:related}

\subsection{Temporal Knowledge Graph Reasoning}

\noindent\textbf{Transductive interpolation (TKG completion).}
TKG completion models predict missing facts at
timestamps within the training range by incorporating
time into the scoring function
\citep{leblay2018deriving,
garcia-duran-etal-2018-learning}, extending tensor
factorisation \citep{tcomplexlacroix2020tensor}, or
applying rotation-based encodings \citep{xu2020tero}.

\noindent\textbf{Transductive extrapolation (TKG forecasting).}
TKG forecasting models predict facts at strictly future
timestamps via GNN-based subgraph evolution
\citep{li2021temporal, li2022tirgn, han2021learning,
sun2021timetraveler, liang2023learn} or rule-based and
explainable reasoning
\citep{liu2022tlogic, gastinger2024history}.

\noindent\textbf{LLM-based inductive TKG reasoning.}
Recent approaches explore inductive TKG inference using
textual descriptions and large language models
\citep{lee2023temporal, liao2024gentkg}, but their
transferability arises from textual semantics rather
than structural representations.

\subsection{KG Foundation Models}

Recent work has moved toward inductive,
vocabulary-agnostic foundation models. NBFNet
\citep{zhu2021neural}, GraIL \citep{teru2020inductive},
and INDIGO \citep{liu2021indigo} introduce
query-conditioned message passing for unseen entities
but assume a fixed relation vocabulary; INGRAM
\citep{lee2023ingram} lifts this by modelling relations
as nodes in a relation graph. \textsc{Ultra}
\citep{galkintowards} and \textsc{Trix}
\citep{zhangtrix} build on this idea, learning
transferable relational structure that lets a
pretrained checkpoint operate across graphs with
disjoint entity and relation vocabularies. FITTER \citep{pan2026fittervocabularyagnosticcrossdomaininference} extends \textsc{Ultra} to TKGs using sinusoidal positional encodings and learned local and global temporal context fusion, with temporal hyperparameters selected for each target graph. In contrast, \Method{} is designed for joint pretraining and cross-TKG transfer: it encodes relative temporal displacement directly in structural messages, and requires no target-graph adaptation.

%% file: sections/preliminaries.tex
\section{Preliminaries}
\label{sec:prelim}

\subsection{Temporal KG Inductive Transfer}
\label{sec:prelim:tkg}

A temporal knowledge graph (TKG) is $G = (V, R, \Tau, Q)$, where $V$, $R$, and $\Tau$ denote the entity, relation, and timestamp sets, respectively, and $Q \subseteq V \times R \times V \times \Tau$ is a set of temporal facts $(s, r, o, \tau)$. We study temporal link prediction, which predicts the missing entity in a query $(s, r, ?, \tau)$ or $(?, r, o, \tau)$.

Our focus is inductive transfer across TKGs, corresponding to the fully-inductive setting in knowledge graph literature. Let $\Gtrain$ and $\Ginf$ denote the TKGs used for training and inference. Following \citet{galkintowards,zhangtrix}, we require their entity, relation, and timestamp vocabularies to be disjoint:
\begin{align*}
  \Vtrain \cap \Vinf &= \emptyset, \quad
  \Rtrain \cap \Rinf = \emptyset, \\
  \Ttrain \cap \Tinf &= \emptyset.
\end{align*}
The pretrained model is applied to $\Ginf$ without fine-tuning. Within the inference TKG, we consider both \emph{interpolation}, where test timestamps lie within the time range of the observed inference graph, and \emph{extrapolation}, where test timestamps occur strictly after it. This setting evaluates whether temporal reasoning learned from source TKGs transfers to entirely unseen entity, relation, and timestamp vocabularies.

%% file: sections/background.tex
\subsection{Base Models}
\label{sec:background}

We build on NBFNet-style \citep{zhu2021neural} KG foundation
models, specifically \textsc{Ultra} \citep{galkintowards} and
\textsc{Trix} \citep{zhangtrix}. Given a query $(s, r, ?)$, both
models derive representations of relations from graph structure
rather than relying on relation-specific embeddings, enabling
transfer to unseen relation vocabularies.

\textsc{Ultra} constructs a relation graph whose nodes correspond
to relations in the original KG and whose edges encode four types
of relation interaction: head-to-head (h2h), head-to-tail (h2t),
tail-to-head (t2h), and tail-to-tail (t2t).
\textsc{Trix} similarly represents relations as nodes, but constructs
and updates its relation graph by jointly refining relation
and entity representations across message-passing layers.
For a query relation $r$, we write
\begin{equation}
  \label{eq:relgnn}
  \{r_{uv}\}
  =
  \operatorname{RelGNN}(G_R \mid r),
\end{equation}
where $G_R$ denotes the model-specific relation graph and
$\operatorname{RelGNN}$ denotes the corresponding relation-level GNN
of \textsc{Ultra} or \textsc{Trix}. The resulting
$r_{uv}$ is the query-conditioned representation of relation
$r_{uv}$ on an edge $(u, r_{uv}, v)$.

Entity reasoning then proceeds by message passing over the original
KG. At layer $\ell$, each edge $(u, r_{uv}, v)$ produces a message
\begin{equation}
  \label{eq:base-message}
  m_{uv}^{(\ell)}
  =
  \phi\!\left(
      h_{u}^{(\ell-1)},
      r_{uv}
  \right),
\end{equation}
where $h_{u}^{(\ell-1)}$ is the representation of the source
entity. Incoming messages are aggregated to update the representation
of $v$:
\begin{equation}
  h_{v}^{(\ell)}
  =
  \operatorname{AGG}
  \left(
      \left\{
      m_{uv}^{(\ell)}
      \mid (u, r_{uv}, v) \in \mathcal{E}
      \right\}
  \right),
\end{equation}
where $\mathcal{E}$ denotes the set of edges in the static KG.

These base models therefore provide transferable structural
representations of entities and relations, but their message functions
do not model when a fact occurs. In the next section, we extend this
entity-side message function to temporal facts with \Method{}, while
leaving the model-specific relation reasoning unchanged.

%% file: sections/method.tex
\section{Method: \Method{}}
\label{sec:method}

We propose \Method{} (Gated Rotary Attention for Temporal
Encoding), a parameter-free mechanism that augments
NBFNet-style KG foundation models with temporal
information. \Cref{fig:grate_mechanism} illustrates the
\Method{} component.

\input{figures/grate_mechanism}

\subsection{Rotary Relative-Time Encoding}
\label{sec:method:rotary}

The base message function defined in
\Cref{eq:base-message},
$m_{uv}^{(\ell)}
= \phi(h_u^{(\ell-1)}, r_{uv})$,
does not incorporate temporal information. Transductive
temporal models commonly learn timestamp-specific
embeddings, which are tied to the timestamps observed during
training and cannot directly represent unseen timestamps.
FITTER \citep{pan2026fittervocabularyagnosticcrossdomaininference}
addresses this limitation by applying sinusoidal encoding to
the positions of timestamps in their ordered sequence,
enabling temporal patterns to transfer across disjoint
timestamp vocabularies. We instead encode each fact by its
temporal displacement from the query. This query-relative
representation directly captures how far the fact occurs from
the query time, which is important for temporal reasoning.

Given a temporal fact $(u,r_{uv},v,t_{uv})$ and a query
$(s,r,?,\tau)$, we define the relative time gap as
\begin{equation}
  \label{eq:relative-time-gap}
  \Delta t_{uv} = \tau - t_{uv}.
\end{equation}
We encode this relative time gap by rotating the message from
\Cref{eq:base-message}:
\begin{equation}
  \label{eq:rotated-message}
  m_{uv}^{\mathrm{rot},(\ell)}
  =
  R(\Delta t_{uv})m_{uv}^{(\ell)},
\end{equation}
where $R(\Delta t)\in\mathbb{R}^{d\times d}$ is a
block-diagonal rotation matrix derived from rotary position
embedding (RoPE) \citep{su2024roformer}. For readability, we
omit the layer superscript below.\footnote{Following
\textsc{Ultra} and \textsc{Trix}, the base model instantiates
$\phi$ in \Cref{eq:base-message} as DistMult:
$m_{uv}=h_u\odot r_{uv}\in\mathbb{R}^{d}$, where $\odot$
denotes the element-wise Hadamard product.}

Concretely, let $d$ denote the even hidden dimension of the
message function. We define the frequency vector as
\begin{equation}
  \label{eq:rope-freq}
  \omega_k
  =
  \frac{1}{10000^{2k/d}},
  \qquad
  k=0,1,\ldots,d/2-1.
\end{equation}
The base $10\,000$ follows the standard RoPE convention
\citep{su2024roformer}. \Cref{app:beta} in Appendix shows that performance
is robust to this choice. The rotation matrix is
block-diagonal:
\begin{equation}
  \label{eq:rope-matrix}
  R(\Delta t)
  =
  \operatorname{diag}\!\left(
  R_0(\Delta t),\ldots,R_{d/2-1}(\Delta t)
  \right),
\end{equation}
where each $2{\times}2$ block rotates the $k$-th coordinate
pair by $\theta_k=\omega_k\Delta t$:
\begin{equation}
  \label{eq:rope-rotation}
  R_k(\Delta t)
  =
  \begin{pmatrix}
    \cos\theta_k & -\sin\theta_k \\
    \sin\theta_k & \phantom{-}\cos\theta_k
  \end{pmatrix}.
\end{equation}
Partitioning a vector $x\in\mathbb{R}^{d}$ into consecutive
pairs $(x_{2k},x_{2k+1})$, the rotation acts independently on
each pair:
\begin{equation}
  \label{eq:rope-pair}
  \begin{aligned}
    \bigl(R(\Delta t)x\bigr)_{2k}
      &=
      x_{2k}\cos\theta_k
      -x_{2k+1}\sin\theta_k,\\
    \bigl(R(\Delta t)x\bigr)_{2k+1}
      &=
      x_{2k}\sin\theta_k
      +x_{2k+1}\cos\theta_k.
  \end{aligned}
\end{equation}

The rotation preserves the norm of the base message
(\Cref{prop:norm-preservation}) while making its orientation
depend on the temporal displacement of the corresponding fact
from the query. Because the
transformation uses relative time gaps and a fixed frequency
basis, it introduces neither timestamp-specific embeddings
nor additional learnable parameters. The same encoding can
therefore be applied to unseen timestamps without learning
new timestamp representations or adapting the temporal
encoding to the target graph. In \Cref{sec:method:gate}, we
use the query to determine how much each rotated message
contributes to predicting the missing entity.

\subsection{Query-Conditioned Gating}
\label{sec:method:gate}

Temporal proximity is an important signal for temporal link
prediction: supporting facts closer to the query time are often more
informative. However, temporal relevance is not determined by
proximity alone. A temporally distant fact may still provide useful
evidence when its structural semantics are strongly aligned with the
query. Temporal reasoning therefore requires accounting for both
temporal displacement and query-dependent semantic relevance.

Existing approaches such as FITTER
\citep{pan2026fittervocabularyagnosticcrossdomaininference}
model temporal locality through local and global temporal contexts,
requiring the temporal neighborhood and their fusion to be configured
for the target TKG. Such choices can become dataset-dependent as
temporal granularity and dynamics vary across TKGs. In contrast, the
relative-time rotation in \Cref{sec:method:rotary} already incorporates
each fact's temporal displacement $\Delta t_{uv}$ into its structural
message. We therefore use the query itself to determine the
contribution of the resulting time-aware message, without constructing
a separate local temporal graph.

Specifically, let $h_{s}^{(\ell-1)}$ denote the
representation of the query head entity at layer $\ell-1$, and let
$r_{q}$ denote the query-relation representation. We define
the query state as
\begin{equation}
  Q_{q} = h_{s}^{(\ell-1)} + r_{q},
\end{equation}
which follows the query-conditioned propagation of the NBFNet-style
base models. Given the rotated message
$m^{\text{rot}}_{uv}$, we compute a scalar gate
\begin{equation}
  \label{eq:gate}
  g_{uv} = \sigma\!\left(\frac{Q_{q}^{\top}\, m^{\text{rot}}_{uv}}{\sqrt{d}}\right),
\end{equation}
where $\sigma$ denotes the sigmoid function and the
$1/\sqrt{d}$ factor stabilizes the magnitude of the dot product.
The final \Method{} message is
\begin{equation}
  \label{eq:grate-message}
  m^{\Method{}}_{uv} = g_{uv} \cdot m^{\text{rot}}_{uv}.
\end{equation}

Because $m^{\text{rot}}_{uv}$ depends on
$\Delta t_{uv}$, its alignment with $Q_{q}$ depends jointly
on the temporal displacement of the supporting fact and its structural
relation to the query. The gate can therefore assign different
contributions to facts at different temporal distances while retaining
temporally distant evidence when it remains relevant to the query.
Unlike approaches that explicitly construct local and global temporal
contexts \citep{pan2026fittervocabularyagnosticcrossdomaininference}, this mechanism does not require a predefined temporal window
or target-specific temporal fusion.

Both the rotation and gating operate on relative time differences and
representations already provided by the base model. \Method{} therefore
introduces no additional learnable parameters (\Cref{app:params}) or
timestamp-specific representations. \Cref{app:theoretical} establishes
the norm preservation, relative-time invariance, and temporal
discriminability of the rotary construction.
\Cref{alg:grate} summarizes the
complete message-passing procedure.

\begin{algorithm}[t]
\small
\caption{\small\Method{} forward pass for layer $\ell$, query
  $(s, r, ?, \tau)$.}
\label{alg:grate}
\begin{algorithmic}[1]
  \REQUIRE node states $\{h_u^{(\ell-1)}\}$,\\
    query relation embedding $r_q$,\\
    edges $u \xrightarrow{r_{uv}} v$ with timestamps $\{t_{uv}\}$,\\
    hidden dimension $d$ (even), RoPE base $b{=}10000$.
  \ENSURE updated node states $\{h_v^{(\ell)}\}$.
  \STATE $\omega_k \gets b^{-2k/d}$ for $k = 0, \ldots, d/2{-}1$
  \STATE $Q_q \gets h_s^{(\ell-1)} + r_q$
    \COMMENT{query state}
  \FORALL{edges $u \xrightarrow{r_{uv}} v$ with timestamp $t_{uv}$}
    \STATE $\Delta t_{uv} \gets \tau - t_{uv}$
    \STATE $m \gets \phi(h_u^{(\ell-1)}, r_{uv})$
      \COMMENT{message}
    \STATE $m^{\mathrm{rot}} \gets R(\Delta t_{uv})\, m$
      via \Cref{eq:rope-pair}
    \STATE $g \gets \sigma\bigl(Q_q^{\top} m^{\mathrm{rot}} / \sqrt{d}\bigr)$
      \COMMENT{gate}
    \STATE $m^{\Method{}}_{uv} \gets g \cdot m^{\mathrm{rot}}$
  \ENDFOR
  \FORALL{nodes $v$}
    \STATE $h_v^{(\ell)} \gets \operatorname{AGG}\bigl(\{m^{\Method{}}_{uv} : (u, r_{uv}, v) \in \mathcal{E}\}\bigr)$
  \ENDFOR
  \STATE \textbf{return} $\{h_v^{(\ell)}\}$
\end{algorithmic}
\end{algorithm}

\subsection{Multi-TKG Temporal Alignment}
\label{sec:method:joint}

Joint pretraining across TKGs requires their temporal coordinate
systems to be comparable. While entities and relations are naturally
disjoint across KGs, \Method{} applies the same rotary
transformation to relative time gaps $\Delta t=\tau-t$. Independently
indexing timestamps within each TKG can therefore assign inconsistent
temporal meaning to the same numerical gap. We instead map all source
timestamps into a unified temporal coordinate system, such that the
same $\Delta t$ represents the same temporal displacement across
pretraining TKGs. This alignment is required only among the source
TKGs during joint pretraining; inference TKGs may contain entirely
unseen timestamps, since \Method{} depends on relative temporal
displacement rather than timestamp-specific parameters.

%% file: figures/grate_mechanism.tex
\begin{figure*}[t]
  \centering
  \includegraphics[width=\textwidth]{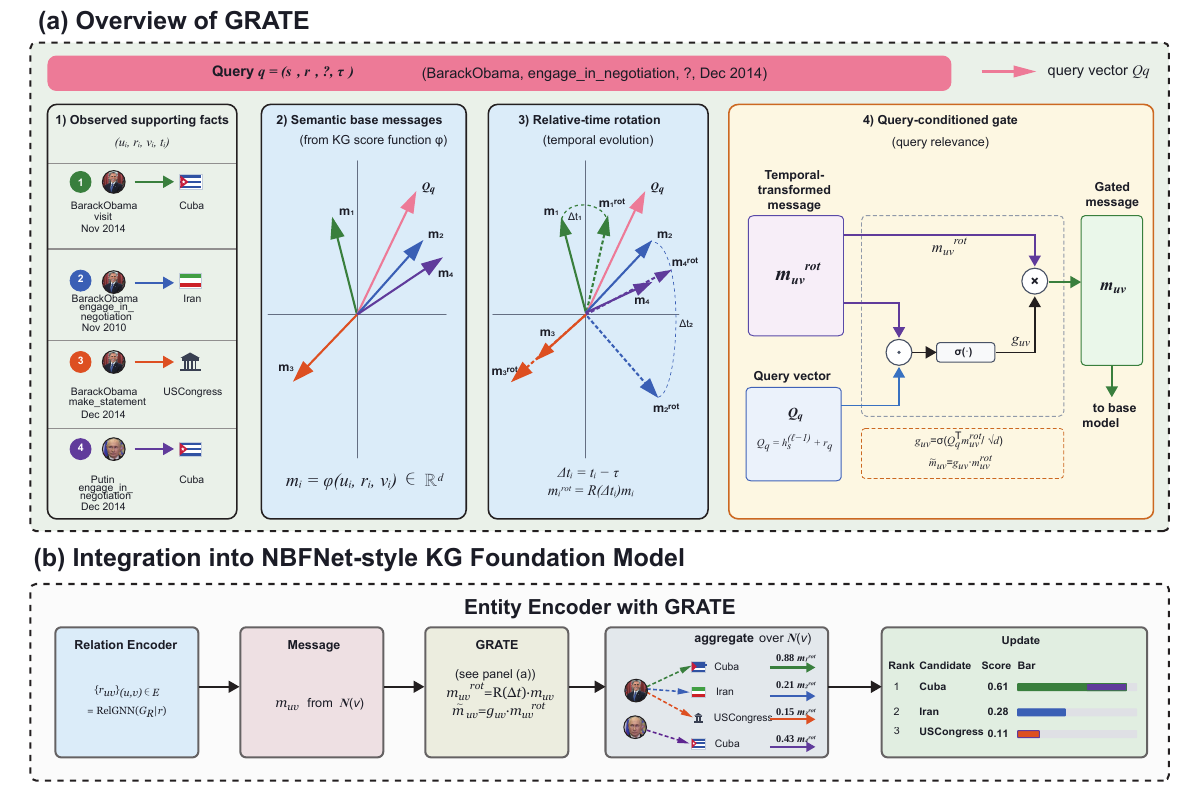}
  \caption{Overview of \Method{} and its integration into NBFNet-style
    KG foundation models.
    (a)~Given a temporal query $(s,r,?,\tau)$, the base model first
    computes structural messages for observed supporting facts.
    \Method{} rotates each message according to its query-relative
    temporal displacement $\Delta t_{uv}=\tau-t_{uv}$, and then applies
    a query-conditioned gate to determine the contribution of each
    time-aware message.
    (b)~\Method{} integrates into the entity-side message passing of the
    base model while leaving its relation reasoning unchanged.}
  \label{fig:grate_mechanism}
\end{figure*}

%% file: sections/benchmark.tex
\section{Experimental Setup}
\label{sec:setup}

\subsection{Evaluation Regimes}
\label{sec:setup:regimes}

We evaluate \Method{} under three regimes:
(1) \emph{pretraining} on ICEWS14 and
ICEWS05-15; (2) \emph{inductive transfer} on the
GDELT and WIKI inductive benchmarks we construct;
(3) \emph{held-out zero-shot transfer} on ICEWS18 and
YAGO under the standard single-step extrapolation
protocol. Regime (2) tests the central claim of
\Method{}; the same jointly pretrained checkpoint is
used in all three regimes.

\subsection{Datasets and Inductive Transfer Benchmark}
\label{sec:setup:datasets}

\paragraph{Pretraining sources.}
We jointly pretrain on ICEWS14 and ICEWS05-15
\citep{garcia-duran-etal-2018-learning,lautenschlager2015icews}, two
political-event TKGs with daily timestamps. We select them
as pretraining sources because, among the six TKGs
considered, they provide the largest relation vocabularies
(230 and 251 relations, respectively), offering a
comparatively diverse set of relational patterns for
pretraining. These are the only TKGs observed during
training.

\paragraph{Inductive transfer benchmark.}
Existing inductive KG benchmarks evaluate transfer across
unseen entities and relations but do not address temporal
graphs. We construct 16 inductive transfer variants from
GDELT \citep{trivedi2017know} and WIKI
\citep{jin2020recurrent} to systematically evaluate
transfer to unseen temporal graphs. Extending the
inductive construction of INGRAM \citep{lee2023ingram} to
temporal quadruples, we vary two orthogonal axes:
$p_{\mathrm{tri}}\in\{0.25,0.50,0.75,1.00\}$, controlling
the fraction of inference-graph edges whose relation and
timestamp types are disjoint from
training,\footnote{V/R/T-disjoint transfer is a deliberate
stress test, most representative of rapidly-evolving
domains where new entities and relations continually appear
over time.} and the temporal mode
$\in\{\textsc{inter},\textsc{extra}\}$, corresponding to
random and chronological message/test splits. Combining
four $p_{\mathrm{tri}}$ settings and two temporal modes for
each of GDELT and WIKI yields $2\times4\times2=16$
benchmark variants. GDELT contains fine-grained daily
event data, whereas WIKI contains yearly encyclopaedic
facts spanning more than two centuries, providing
contrasting temporal settings for transfer evaluation.
Full construction details are in \Cref{app:datasets}.

\paragraph{Held-out forecasting.}
ICEWS18 (daily political events) and YAGO (yearly
encyclopaedic facts), both from \citet{jin2020recurrent},
are evaluated under the standard single-step extrapolation
protocol, where the full training set of the target TKG
serves as the inference graph for both validation and test
queries. In contrast, the inductive transfer benchmarks
above use a stricter protocol: the inference graph is a
disjoint subset of an unseen graph, and test queries are
answered from this partial observed context alone.

\paragraph{Dataset statistics.}
Per-dataset statistics for all six source TKGs are in
\Cref{tab:source_datasets} (\Cref{app:datasets});
per-variant statistics for the 16 inductive benchmarks
are in \Cref{tab:gdelt_sweep,tab:wiki_sweep}.
\Cref{app:representativeness} verifies that the
constructed splits preserve the structural properties of
their source TKGs.

\subsection{Evaluation Protocol and Baselines}
\label{sec:benchmark:protocol}

\paragraph{Metrics.}
We report MRR and Hits@1/10 averaged over both tail prediction
$(s,r,?,\tau)$ and head prediction $(?,r,o,\tau)$, using the
time-aware filtered setting of \citet{goel2020diachronic}.

\paragraph{Baselines.}
We compare against baselines appropriate to each evaluation regime.
(i)~\emph{Pretraining}: \textsc{Ultra}
\citep{galkintowards} and \textsc{Trix} \citep{zhangtrix},
jointly pretrained on the same ICEWS14/ICEWS05-15 mixture without
\Method{}, isolate the contribution of the temporal mechanism.
We additionally report FITTER
\citep{pan2026fittervocabularyagnosticcrossdomaininference},
trained and tuned separately on each dataset.
A broader comparison with transductive interpolation models is
provided in \Cref{app:transductive_full}.
(ii)~\emph{Inductive transfer}: we compare against the count-based
recurrency predictor of \citet{gastinger2024history} and FITTER using
its ICEWS14-trained checkpoint with per-target hyperparameter
selection. Most existing temporal models cannot directly operate
under the V/R/T-disjoint setting; we discuss their applicability in
\Cref{app:baseline_inapplicability}.
(iii)~\emph{Held-out forecasting}: we compare against FITTER using
its ICEWS14-trained checkpoint, ICL \citep{lee2023temporal}, and
GenTKG \citep{liao2024gentkg}. A broader comparison with
transductive forecasting models is provided in
\Cref{app:transductive_heldout}.

\paragraph{Implementation.}
The same jointly pretrained checkpoint is used for all \Method{}
evaluations without target-specific fine-tuning. Hyperparameters,
optimization, and hardware details are provided in
\Cref{app:implementation}. Zero-shot evaluation is deterministic,
so we report single-run results.\footnote{Code and benchmarks:
\url{https://anonymous.4open.science/r/GRATE-F431/}.}

%% file: sections/experiments.tex
\section{Results}
\label{sec:exp:analysis}

We evaluate a single \Method{} checkpoint jointly pretrained on ICEWS14 and ICEWS05-15. \Cref{sec:exp:transductive} reports pretraining performance on ICEWS14 and ICEWS05-15. \Cref{sec:exp:zeroshot} evaluates inductive transfer performance across the 16 benchmarks constructed in \Cref{sec:setup:datasets}. \Cref{sec:exp:heldout} tests held-out forecasting on ICEWS18 and YAGO. \Cref{sec:exp:ablations} isolates the contribution of each \Method{} component.

\subsection{Pretraining Performance}
\label{sec:exp:transductive}

\begin{table}[t]
  \centering
  \scriptsize
  \setlength{\tabcolsep}{2pt}
  \resizebox{\columnwidth}{!}{%
  \begin{tabular}{l ccc ccc}
    \toprule
    & \multicolumn{3}{c}{ICEWS14} & \multicolumn{3}{c}{ICEWS05-15} \\
    \cmidrule(lr){2-4} \cmidrule(lr){5-7}
    Method & MRR & H@1 & H@10 & MRR & H@1 & H@10 \\
    \midrule
    FITTER          & .583 & ---   & .760 & .590 & ---   & .787 \\
    \midrule
    \textsc{Ultra}              & .495 & .373 & .731 & .451 & .316 & .720 \\
    $+$\,\Method{} & \textbf{.620} & \textbf{.520} & \textbf{.793} & \textbf{.621} & .500 & .819 \\
    $\%\,\Delta$                     & $+25.3\%$ & $+39.4\%$ & $+8.5\%$ & $+37.7\%$ & $+58.2\%$ & $+13.8\%$ \\
    \midrule
    \textsc{Trix}               & .486 & .366 & .719 & .456 & .322 & .722 \\
    $+$\,\Method{} & .609 & .507 & \textbf{.793} & .617 & \textbf{.501} & \textbf{.830} \\
    $\%\,\Delta$                     & $+25.3\%$ & $+38.5\%$ & $+10.3\%$ & $+35.3\%$ & $+55.6\%$ & $+15.0\%$ \\
    \bottomrule
  \end{tabular}}
  \caption{Link prediction result on the pretraining TKGs. FITTER is trained and tuned separately on
    each dataset; all other methods use a single jointly
    pretrained checkpoint. $\%\,\Delta$ rows show relative
    improvement over the base model without \Method{}.
    Full comparison with transductive baselines in
    \Cref{app:transductive_full}.
    Best per column \textbf{bold}.}
  \label{tab:transductive}
\end{table}

We evaluate the jointly pretrained checkpoint on
ICEWS14 and ICEWS05-15, the two TKGs used during
pretraining. Adding \Method{} substantially improves both
\textsc{Ultra} and \textsc{Trix} across all metrics. On
ICEWS14, \Method{} increases MRR from .495 to .620 for
\textsc{Ultra} and from .486 to .609 for \textsc{Trix}.
Similar improvements appear on ICEWS05-15, where MRR
increases from .451 to .621 and from .456 to .617,
respectively. The improvements are particularly pronounced
for Hits@1, with relative gains between 38\% and 58\%,
showing that temporal information substantially improves
top-ranked predictions over the base models.

These gains are consistent across both base models and both
pretraining TKGs, indicating that the benefit is not
specific to one model architecture. Moreover, a single
jointly pretrained \Method{} checkpoint reaches or exceeds
FITTER, which is trained and tuned separately for each
dataset. This establishes that joint pretraining does not
sacrifice in-distribution temporal reasoning performance.

\subsection{Inductive transfer}
\label{sec:exp:zeroshot}

\begin{table}[t]
  \centering
  \scriptsize
  \setlength{\tabcolsep}{2.5pt}
  \begin{tabular}{l ccccc}
    \toprule
    & \multicolumn{2}{c}{\textsc{GDELTIndT}} & \multicolumn{2}{c}{\textsc{WIKIIndT}} & \\
    \cmidrule(lr){2-3} \cmidrule(lr){4-5}
    Method & Inter.\ (4) & Extra.\ (4) & Inter.\ (4) & Extra.\ (4) & Avg.\ (16) \\
    \midrule
    Recurrency          & .225 & .158 & .556 & .532 & .368 \\
    FITTER              & .267 & .273 & .804 & .958 & .575 \\
    \midrule
    \textsc{Ultra}      & .216 & .255 & .776 & .923 & .542 \\
    $+$\,\Method{}      & .249 & .291 & .781 & .961 & .570 \\
    \midrule
    \textsc{Trix}       & .228 & .221 & .801 & .899 & .537 \\
    $+$\,\Method{}      & \textbf{.412} & \textbf{.315} & \textbf{.842} & \textbf{.968} & \textbf{.634} \\
    \bottomrule
  \end{tabular}
  \caption{Inductive transfer MRR averaged over the four
    $p_{\mathrm{tri}}$ settings per dataset and mode.
    All methods except FITTER use the same jointly
    pretrained checkpoint. FITTER uses the ICEWS14
    checkpoint from the original paper.
    Best per column \textbf{bold}.
    Complete per-$p_{\mathrm{tri}}$ results in
    \Cref{app:sweep_full}.}
  \label{tab:transfer_summary}
\end{table}

\begin{figure}[t]
  \centering
  \includegraphics[width=\columnwidth]{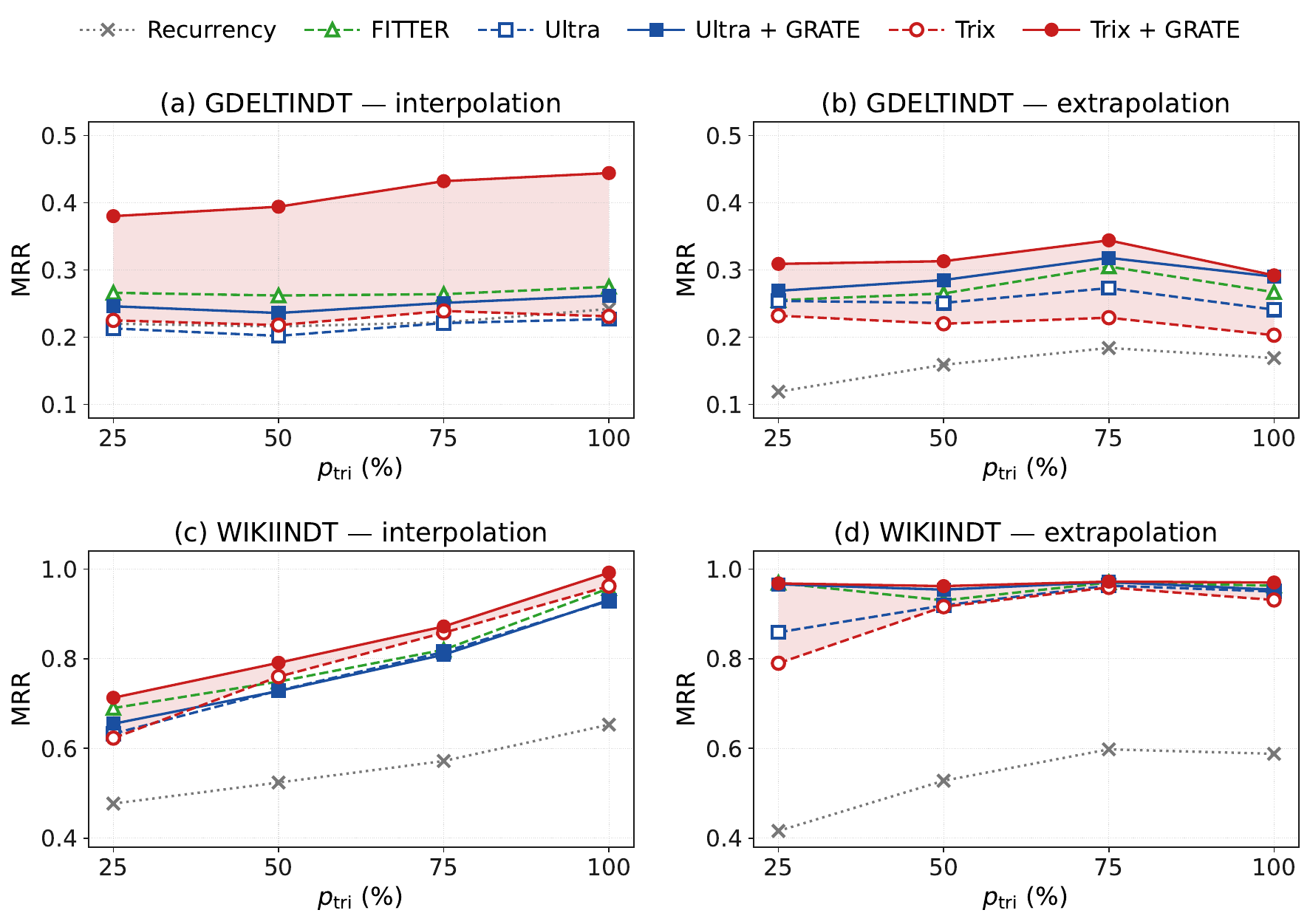}
  \caption{Zero-shot inductive transfer MRR on
    \textsc{GDELTIndT} and \textsc{WIKIIndT} as a function
    of $p_{\mathrm{tri}}$.
    Hits@$1$/Hits@$10$ variants are in
    \Cref{fig:sweep_results_h1,fig:sweep_results_h10} and the full table is in
    \Cref{app:sweep_full}.}
  \label{fig:sweep_results}
\end{figure}

\Cref{tab:transfer_summary} summarizes performance across
the 16 inductive transfer benchmarks constructed
from GDELT and WIKI, covering different levels of vocabulary
overlap between training and inference.
\Cref{fig:sweep_results} shows MRR as a function of
$p_{\mathrm{tri}}$; complete numerical
results are provided in \Cref{app:sweep_full}. The same
checkpoint jointly pretrained on ICEWS14 and ICEWS05-15 is
evaluated throughout without target-specific fine-tuning.
Averaged over all 16 settings, \Method{} improves
\textsc{Ultra} from .542 to .570 MRR and \textsc{Trix} from
.537 to .634. \textsc{Trix}\,+\,\Method{} achieves the best
overall result, outperforming FITTER (.575) and the
recurrence baseline (.368). The two benchmark suites cover
markedly different temporal conditions. \textsc{GDELTIndT} captures
fine-grained event dynamics where
recurrence is substantially less predictive (.225/.158 MRR
under interpolation/extrapolation) than on \textsc{WIKIIndT}
(.556/.532), which is organized at yearly resolution over
more than two centuries. On \textsc{GDELTIndT}, \Method{} substantially
improves \textsc{Trix} from .228 to .412 under interpolation
and from .221 to .315 under extrapolation. On \textsc{WIKIIndT},
\Method{} improves \textsc{Trix} from .899 to .968 under
extrapolation and \textsc{Ultra} from .923 to .961.
Interestingly, \Cref{fig:sweep_results} reveals contrasting
trends across $p_{\mathrm{tri}}$: \Method{}'s gains on \textsc{GDELTIndT}
generally increase with $p_{\mathrm{tri}}$, whereas on \textsc{WIKIIndT}
they are largest at lower $p_{\mathrm{tri}}$ and narrow as
$p_{\mathrm{tri}}$ increases. Thus, \Method{} remains
effective despite substantial differences in both recurrence
strength and inductive
difficulty.

\subsection{Zero-shot transfer}
\label{sec:exp:heldout}

\Cref{tab:heldout} reports zero-shot transfer to ICEWS18
and YAGO under the single-step extrapolation protocol,
using the same jointly pretrained checkpoint without
fine-tuning. We compare against the LLM-based forecasters
ICL \citep{lee2023temporal} and GenTKG
\citep{liao2024gentkg}; MRR is omitted for these methods
because they decode candidates rather than rank all
entities. A full comparison with transductive baselines is
provided in \Cref{app:transductive_heldout}.

\begin{table}[t]
  \centering
  \scriptsize
  \setlength{\tabcolsep}{2.5pt}
  \begin{tabular}{l ccc ccc}
    \toprule
    & \multicolumn{3}{c}{ICEWS18} & \multicolumn{3}{c}{YAGO} \\
    \cmidrule(lr){2-4} \cmidrule(lr){5-7}
    Method & MRR & H@1 & H@10 & MRR & H@1 & H@10 \\
    \midrule
    ICL      & --- & .182 & .414 & --- & .726 & .846 \\
    GenTKG   & --- & \textbf{.243} & .421 & --- & .792 & .843 \\
    \midrule
    FITTER   & .228 & .128 & .444 & .792 & .736 & .886 \\
    \midrule
    \textsc{Ultra}                           & .180 & .092 & .369 & .726 & .647 & .872 \\
    $+$\,\Method{}             & .227 & .126 & .447 & .793 & .738 & .890 \\
    $\%\,\Delta$                             & $+26.1\%$ & $+37.0\%$ & $+21.1\%$ & $+9.2\%$ & $+14.1\%$ & $+2.1\%$ \\
    \midrule
    \textsc{Trix}                            & \textbf{.274} & .170 & \textbf{.484} & .800 & .740 & \textbf{.916} \\
     $+$\,\Method{}              & .273 & .179 & .461 & \textbf{.858} & \textbf{.822} & .905 \\
    $\%\,\Delta$                             & $-0.4\%$ & $+5.3\%$ & $-4.8\%$ & $+7.3\%$ & $+11.1\%$ & $-1.2\%$ \\
    \bottomrule
  \end{tabular}
  \caption{Zero-shot transfer to ICEWS18 and YAGO. FITTER
    uses a checkpoint pretrained on ICEWS14 as in the
    original paper; all other structure-based methods use
    the same jointly pretrained checkpoint. A full comparison
    with transductive baselines is in
    \Cref{app:transductive_heldout}.}
  \label{tab:heldout}
\end{table}

Despite relying purely on graph structure without textual
descriptions or language-model priors, \Method{} is
competitive with the LLM-based forecasters on both held-out
datasets. On ICEWS18, \textsc{Trix}\,+\,\Method{} achieves
higher Hits@$10$ than both ICL and GenTKG, while GenTKG
obtains higher Hits@$1$ partly through entity-relation
co-occurrence filtering that narrows its candidate space;
\Method{} instead ranks over all candidate entities. On
YAGO, \textsc{Trix}\,+\,\Method{} achieves .858 MRR and
.822 Hits@$1$ and surpasses FITTER on both metrics. \Method{}'s gains are larger when the base model has more room to improve: \textsc{Ultra}\,+\,\Method{} on ICEWS18 closes the gap to FITTER, while \textsc{Trix} already performs strongly on its own. These
results show that the jointly
pretrained model transfers to unseen temporal graphs without
target-specific fine-tuning.

\subsection{Component ablation}
\label{sec:exp:ablations}

\Cref{tab:ablations} isolates the contribution of each
\Method{} component by incrementally augmenting the
\textsc{Trix} base model. All variants share the same
architecture and training setup, differing only in the
entity-side message function.

\begin{table}[t]
  \centering
  \small
  \setlength{\tabcolsep}{2pt}
  \begin{tabular}{lcc}
    \toprule
                  & ICEWS14         & \textsc{GDELT-100i} \\
    Configuration & (transd.\ MRR)  & (zs MRR) \\
    \midrule
    \textsc{Trix} (distmult)                  & .486          & .231          \\
    \quad $+\,m \odot \mathrm{SinCos}(\Delta t)$  & .560          & .320          \\
    \quad + rotation                          & .589          & .377          \\
    \midrule
    \quad rotation + rel.\ gate               & .603          & .410          \\
    $+$\,\Method{}                             & \textbf{.609} & \textbf{.444} \\
    \bottomrule
  \end{tabular}
  \caption{Component ablation of \Method{} on \textsc{Trix}.
    \textsc{GDELT-100i} =
    \textsc{GDELTIndT-100-inter} (zero-shot).}
  \label{tab:ablations}
\end{table}

Replacing the static DistMult message with its rotated
form $R(\Delta t)$ yields gains of $+0.103$ MRR on
ICEWS14 and $+0.146$ MRR on
\textsc{GDELTIndT-100-inter} without adding parameters,
showing that most of the temporal signal can be captured
by relative-time alignment alone. Rotation also
outperforms an element-wise sinusoidal alternative
$m \odot \mathrm{SinCos}(\Delta t)$ by $0.029$ / $0.057$
MRR, since rotation preserves the message norm whereas
sinusoidal modulation rescales each dimension
independently. Adding a relation-conditioned gate on top
of rotation yields further gains of $+0.014$
MRR transductively and $+0.033$ MRR under inductive transfer
evaluation; the effect is larger in the zero-shot setting,
where structural and recurrent cues are weaker. Switching to query-conditioned gating yields modest
in-distribution gains ($+0.006$ MRR) and larger
inductive transfer gains ($+0.034$ MRR), confirming the
value of query-aware edge selection.
Each step in \Cref{tab:ablations} adds a single design
choice; removing any one degrades performance, and the
gap widens under the strictest inductive transfer regime,
indicating that the four design choices function
together rather than additively.

%% file: sections/conclusion.tex
\section{Conclusion}
\label{sec:conclusion}

We introduced \Method{}, a parameter-free temporal message
function that extends KG foundation models with temporal
reasoning by combining query-relative rotary encoding with
query-conditioned gating. We further constructed 16 inductive
transfer benchmarks to systematically evaluate temporal
reasoning across different levels of vocabulary overlap.
\Method{} enables joint multi-TKG pretraining and transfers 
without target-specific hyperparameter
selection, achieving strong temporal link prediction on proposed datasets. Our results show that transferable temporal reasoning
can be incorporated into KG foundation models through a
lightweight, modular mechanism.

%% file: sections/limitations.tex
\section*{Limitations}
\label{app:limitations}

\Method{} encodes time purely through relative index
differences and does not explicitly model temporal
granularity or irregular intervals, which weakens its
effect on coarse-grained datasets. In addition, while
\Method{} adds no parameters, the underlying NBFNet-style
message passing remains computationally expensive on
large TKGs. Combining relative-time encoding with more
scalable propagation strategies, and broader pretraining
mixes including GDELT or interval-style TKGs, is a
natural direction for future work. The current pretraining sources share daily granularity;
extending to mixed-granularity pretraining is left to future work.

%% file: sections/app_sweep_full.tex
\section{Full Numerical Results}
\label{app:sweep_full}

\subsection{Inductive Transfer Benchmark Tables}

\begin{table*}[!tp]
  \centering
  \scriptsize
  \setlength{\tabcolsep}{3pt}

  \subtabtitle{(a) \textsc{GDELTIndT}: interpolation mode}

  \begin{tabular}{l ccc ccc ccc ccc}
    \toprule
    & \multicolumn{3}{c}{$p_{25}$} & \multicolumn{3}{c}{$p_{50}$} & \multicolumn{3}{c}{$p_{75}$} & \multicolumn{3}{c}{$p_{100}$} \\
    \cmidrule(lr){2-4} \cmidrule(lr){5-7} \cmidrule(lr){8-10} \cmidrule(lr){11-13}
    Method & MRR & H@1 & H@10 & MRR & H@1 & H@10 & MRR & H@1 & H@10 & MRR & H@1 & H@10 \\
    \midrule
    Recurrency               & .220 & .000 & .502 & .216 & .000 & .495 & .222 & .000 & .492 & .242 & .000 & .537 \\
    \midrule
    FITTER                   & .266 & .173 & .448 & .262 & .171 & .435 & .264 & .178 & .425 & .275 & .187 & .442 \\
    \midrule
    \textsc{Ultra}           & .213 & .126 & .383 & .202 & .114 & .372 & .221 & .135 & .386 & .227 & .130 & .416 \\
    $+$\,\Method{}& .246 & .150 & .435 & .236 & .139 & .427 & .251 & .158 & .434 & .262 & .162 & .458 \\
    $\Delta$ (\textsc{Ultra})    & $+.033$ & $+.024$ & $+.052$ & $+.034$ & $+.025$ & $+.055$ & $+.030$ & $+.023$ & $+.048$ & $+.035$ & $+.032$ & $+.042$ \\
    \midrule
    \textsc{Trix}            & .225 & .134 & .407 & .218 & .123 & .403 & .239 & .138 & .433 & .231 & .126 & .440 \\
    $+$\,\Method{}& \textbf{.380} & \textbf{.263} & \textbf{.615} & \textbf{.394} & \textbf{.275} & \textbf{.636} & \textbf{.432} & \textbf{.312} & \textbf{.671} & \textbf{.444} & \textbf{.310} & \textbf{.723} \\
    $\Delta$ (\textsc{Trix})     & $+.155$ & $+.129$ & $+.208$ & $+.176$ & $+.152$ & $+.233$ & $+.193$ & $+.174$ & $+.238$ & $+.213$ & $+.184$ & $+.283$ \\
    \bottomrule
  \end{tabular}
  \subtabgap
  \subtabtitle{(b) \textsc{GDELTIndT}: extrapolation mode}

  \begin{tabular}{l ccc ccc ccc ccc}
    \toprule
    & \multicolumn{3}{c}{$p_{25}$} & \multicolumn{3}{c}{$p_{50}$} & \multicolumn{3}{c}{$p_{75}$} & \multicolumn{3}{c}{$p_{100}$} \\
    \cmidrule(lr){2-4} \cmidrule(lr){5-7} \cmidrule(lr){8-10} \cmidrule(lr){11-13}
    Method & MRR & H@1 & H@10 & MRR & H@1 & H@10 & MRR & H@1 & H@10 & MRR & H@1 & H@10 \\
    \midrule
    Recurrency               & .119 & .000 & .272 & .159 & .000 & .358 & .184 & .000 & .401 & .169 & .000 & .369 \\
    \midrule
    FITTER                   & .255 & .164 & .428 & .265 & .178 & .428 & .305 & .220 & .462 & .267 & .177 & .444 \\
    \midrule
    \textsc{Ultra}           & .254 & .158 & .444 & .251 & .152 & .441 & .273 & .173 & .474 & .241 & .143 & .432 \\
    $+$\,\Method{}& .269 & .167 & .467 & .285 & .181 & .488 & .318 & .214 & .523 & .290 & \textbf{.189} & .491 \\
    $\Delta$ (\textsc{Ultra})    & $+.015$ & $+.009$ & $+.023$ & $+.034$ & $+.029$ & $+.047$ & $+.045$ & $+.041$ & $+.049$ & $+.049$ & $+.046$ & $+.059$ \\
    \midrule
    \textsc{Trix}            & .232 & .139 & .414 & .220 & .128 & .395 & .229 & .130 & .421 & .203 & .112 & .379 \\
    $+$\,\Method{}& \textbf{.309} & \textbf{.204} & \textbf{.514} & \textbf{.313} & \textbf{.206} & \textbf{.516} & \textbf{.344} & \textbf{.237} & \textbf{.547} & \textbf{.292} & .186 & \textbf{.503} \\
    $\Delta$ (\textsc{Trix})     & $+.077$ & $+.065$ & $+.100$ & $+.093$ & $+.078$ & $+.121$ & $+.115$ & $+.107$ & $+.126$ & $+.089$ & $+.074$ & $+.124$ \\
    \bottomrule
  \end{tabular}
  \subtabgap
  \subtabtitle{(c) \textsc{WIKIIndT}: interpolation mode}

  \begin{tabular}{l ccc ccc ccc ccc}
    \toprule
    & \multicolumn{3}{c}{$p_{25}$} & \multicolumn{3}{c}{$p_{50}$} & \multicolumn{3}{c}{$p_{75}$} & \multicolumn{3}{c}{$p_{100}$} \\
    \cmidrule(lr){2-4} \cmidrule(lr){5-7} \cmidrule(lr){8-10} \cmidrule(lr){11-13}
    Method & MRR & H@1 & H@10 & MRR & H@1 & H@10 & MRR & H@1 & H@10 & MRR & H@1 & H@10 \\
    \midrule
    Recurrency               & .477 & .000 & .813 & .524 & .000 & .855 & .572 & .000 & .888 & .653 & .000 & .993 \\
    \midrule
    FITTER                   & .690 & .630 & .798 & .749 & .708 & .823 & .820 & .797 & .859 & .957 & .955 & .961 \\
    \midrule
    \textsc{Ultra}           & .633 & .524 & .797 & .729 & .666 & .829 & .815 & .784 & .868 & .929 & .907 & .960 \\
    $+$\,\Method{}& .655 & .566 & .795 & .728 & .666 & .822 & .809 & .776 & .866 & .931 & .912 & .960 \\
    $\Delta$ (\textsc{Ultra})    & $+.022$ & $+.042$ & $-.002$ & $-.001$ & $\pm.000$ & $-.007$ & $-.006$ & $-.008$ & $-.002$ & $+.002$ & $+.005$ & $\pm.000$ \\
    \midrule
    \textsc{Trix}            & .623 & .505 & \textbf{.819} & .760 & .691 & \textbf{.860} & .858 & .826 & .899 & .962 & .937 & \textbf{.996} \\
    $+$\,\Method{}& \textbf{.713} & \textbf{.652} & \textbf{.819} & \textbf{.791} & \textbf{.749} & \textbf{.860} & \textbf{.872} & \textbf{.848} & \textbf{.912} & \textbf{.992} & \textbf{.991} & .995 \\
    $\Delta$ (\textsc{Trix})     & $+.090$ & $+.147$ & $\pm.000$ & $+.031$ & $+.058$ & $\pm.000$ & $+.014$ & $+.022$ & $+.013$ & $+.030$ & $+.054$ & $-.001$ \\
    \bottomrule
  \end{tabular}
  \subtabgap
  \subtabtitle{(d) \textsc{WIKIIndT}: extrapolation mode}

  \begin{tabular}{l ccc ccc ccc ccc}
    \toprule
    & \multicolumn{3}{c}{$p_{25}$} & \multicolumn{3}{c}{$p_{50}$} & \multicolumn{3}{c}{$p_{75}$} & \multicolumn{3}{c}{$p_{100}$} \\
    \cmidrule(lr){2-4} \cmidrule(lr){5-7} \cmidrule(lr){8-10} \cmidrule(lr){11-13}
    Method & MRR & H@1 & H@10 & MRR & H@1 & H@10 & MRR & H@1 & H@10 & MRR & H@1 & H@10 \\
    \midrule
    Recurrency               & .416 & .000 & .624 & .528 & .000 & .794 & .598 & .000 & .903 & .588 & .000 & .891 \\
    \midrule
    FITTER                   & \textbf{.969} & \textbf{.965} & .975 & .930 & .917 & .964 & .970 & .965 & .980 & .963 & .955 & .973 \\
    \midrule
    \textsc{Ultra}           & .859 & .744 & .979 & .919 & .874 & \textbf{.972} & .963 & .946 & \textbf{.984} & .950 & .930 & .975 \\
    $+$\,\Method{}& .966 & .954 & .980 & .954 & .941 & .971 & .971 & .959 & .983 & .954 & .933 & \textbf{.979} \\
    $\Delta$ (\textsc{Ultra})    & $+.107$ & $+.210$ & $+.001$ & $+.035$ & $+.067$ & $-.001$ & $+.008$ & $+.013$ & $-.001$ & $+.004$ & $+.003$ & $+.004$ \\
    \midrule
    \textsc{Trix}            & .790 & .671 & .980 & .916 & .867 & .971 & .959 & .945 & .978 & .931 & .895 & .974 \\
    $+$\,\Method{}& .968 & .959 & \textbf{.981} & \textbf{.962} & \textbf{.955} & \textbf{.972} & \textbf{.972} & \textbf{.966} & .980 & \textbf{.970} & \textbf{.963} & .977 \\
    $\Delta$ (\textsc{Trix})     & $+.178$ & $+.288$ & $+.001$ & $+.046$ & $+.088$ & $+.001$ & $+.013$ & $+.021$ & $+.002$ & $+.039$ & $+.068$ & $+.003$ \\
    \bottomrule
  \end{tabular}
  \caption{Zero-shot inductive transfer to the
    \textsc{GDELTIndT} and \textsc{WIKIIndT} benchmark
    suites under the time-aware filter. Each
    $p_{\mathrm{tri}}$ column
    reports MRR / H@1 / H@10. Rows grouped by base model, with
    $\Delta$ rows showing the
    \Method{}\,$-$\,no-\Method{} difference per cell. All
    learned methods use the same joint checkpoint pretrained
    on ICEWS14\,+\,ICEWS05-15; the \textsc{Ultra} and
    \textsc{Trix} rows are retrained on the same temporal
    data without the \Method{} module.
    FITTER \citep{pan2026fittervocabularyagnosticcrossdomaininference}
    uses the ICEWS14 checkpoint from the original paper.
    Recurrency baseline
    from \citet{gastinger2024history}; \textsc{Ultra} from
    \citet{galkintowards}; \textsc{Trix} from
    \citet{zhangtrix}. Best per column \textbf{bold}.}
  \label{tab:sweep_results}
\end{table*}

\Cref{tab:sweep_results} reports the full numerical
results underlying \Cref{fig:sweep_results}, including
Hits@$1$ and Hits@$10$ alongside MRR for every cell of
the \textsc{GDELTIndT} and \textsc{WIKIIndT} benchmark
suites. We highlight four cross-cutting observations.

\topic{GDELT interpolation (Table~\ref{tab:sweep_results}a).}
\textsc{Trix}\,+\,\Method{} achieves the strongest results
across all $p_{\mathrm{tri}}$ levels, with gains over the
static \textsc{Trix} base model growing monotonically from
$+0.155$ MRR at $p_{25}$ to $+0.213$ MRR at $p_{100}$.
This monotone increase confirms that \Method{}'s gate
becomes more valuable as the inference graph contains a
larger fraction of relation- and timestamp-disjoint edges,
where direct recurrence matching is less reliable.
\textsc{Ultra}\,+\,\Method{} also improves consistently
($+0.030$ to $+0.035$ MRR) but at a smaller scale,
consistent with the relation-graph difference discussed in
\Cref{sec:exp:zeroshot}. FITTER, evaluated zero-shot from
its ICEWS14 checkpoint, outperforms both static base models
and sits between \textsc{Ultra}\,+\,\Method{} and
\textsc{Trix}\,+\,\Method{}.

\topic{GDELT extrapolation (Table~\ref{tab:sweep_results}b).}
Gains persist under the stricter extrapolation split but
are smaller than under interpolation, consistent with
GDELT's dense and diverse event patterns being harder to
generalise across the temporal boundary.
\textsc{Trix}\,+\,\Method{} maintains consistent positive
gains ($+0.077$ to $+0.115$ MRR), peaking at $p_{75}$
where the combination of moderate disjointness and
sufficient temporal context maximises the gate's leverage.
\textsc{Ultra}\,+\,\Method{} likewise improves at every
$p_{\mathrm{tri}}$ level ($+0.015$ to $+0.049$ MRR),
with gains increasing alongside $p_{\mathrm{tri}}$.

\topic{WIKI interpolation (Table~\ref{tab:sweep_results}c).}
WIKI is highly recurrent and static base models already
perform strongly, yet \textsc{Trix}\,+\,\Method{} improves
over static \textsc{Trix} at every $p_{\mathrm{tri}}$ level
($+0.014$ to $+0.090$ MRR). The largest gain appears at
$p_{25}$ ($+0.090$ MRR), where the inference graph is
sparsest and the temporal gate compensates for limited
structural evidence. At higher $p_{\mathrm{tri}}$, the
abundant recurrent structure already captured by the static
base model leaves less room for temporal gating, yet
\Method{} still provides consistent positive gains.
\textsc{Ultra}\,+\,\Method{} is near-parity with static
\textsc{Ultra} on this dense, recurrent source, consistent
with the pattern discussed in \Cref{sec:exp:zeroshot}.

\topic{WIKI extrapolation (Table~\ref{tab:sweep_results}d).}
\textsc{Trix}\,+\,\Method{} improves over static
\textsc{Trix} on the link prediction task at every $p_{\mathrm{tri}}$ level ($+0.013$
to $+0.178$ MRR). The largest gain appears at $p_{25}$
($+0.178$ MRR, $+0.288$ Hits@$1$), where the static base
model has the most room to improve; at higher
$p_{\mathrm{tri}}$ the recurrent structure already provides
strong cues and the marginal gain narrows.
\textsc{Ultra}\,+\,\Method{} follows the same pattern,
with a large gain at $p_{25}$ ($+0.107$ MRR) that tapers
as the static base model strengthens at higher
$p_{\mathrm{tri}}$.

\subsection{Additional Hits@K Results}

\begin{figure*}[!tp]
  \centering
  \includegraphics[width=0.95\textwidth]{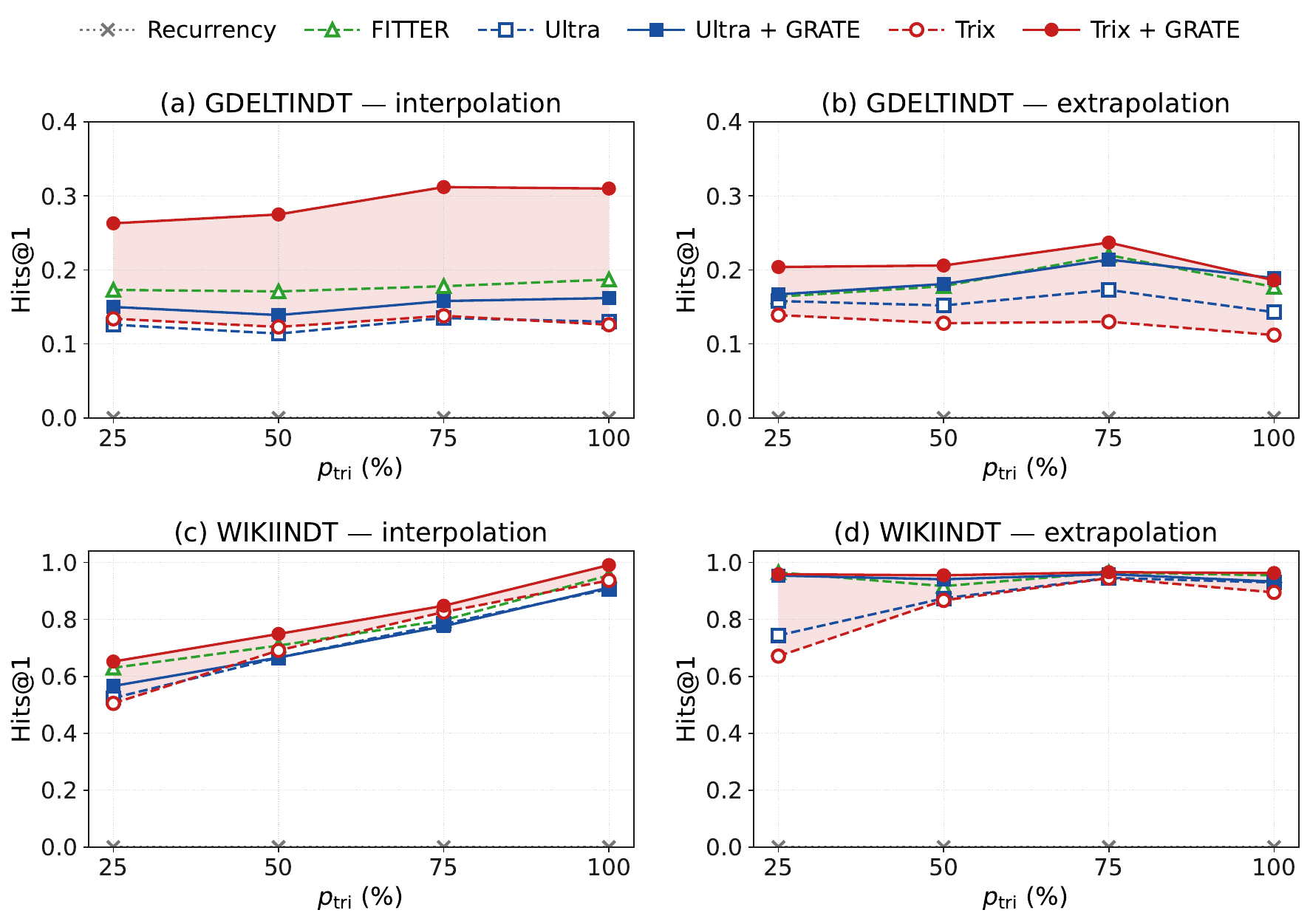}
  \caption{Hits@$1$ counterpart of \Cref{fig:sweep_results}.
    The recurrency baseline sits at $0$ across all panels because the
    count-based predictor breaks ties uniformly and never returns the
    gold tail at exact rank $1$.}
  \label{fig:sweep_results_h1}
\end{figure*}

\begin{figure*}[!tp]
  \centering
  \includegraphics[width=0.95\textwidth]{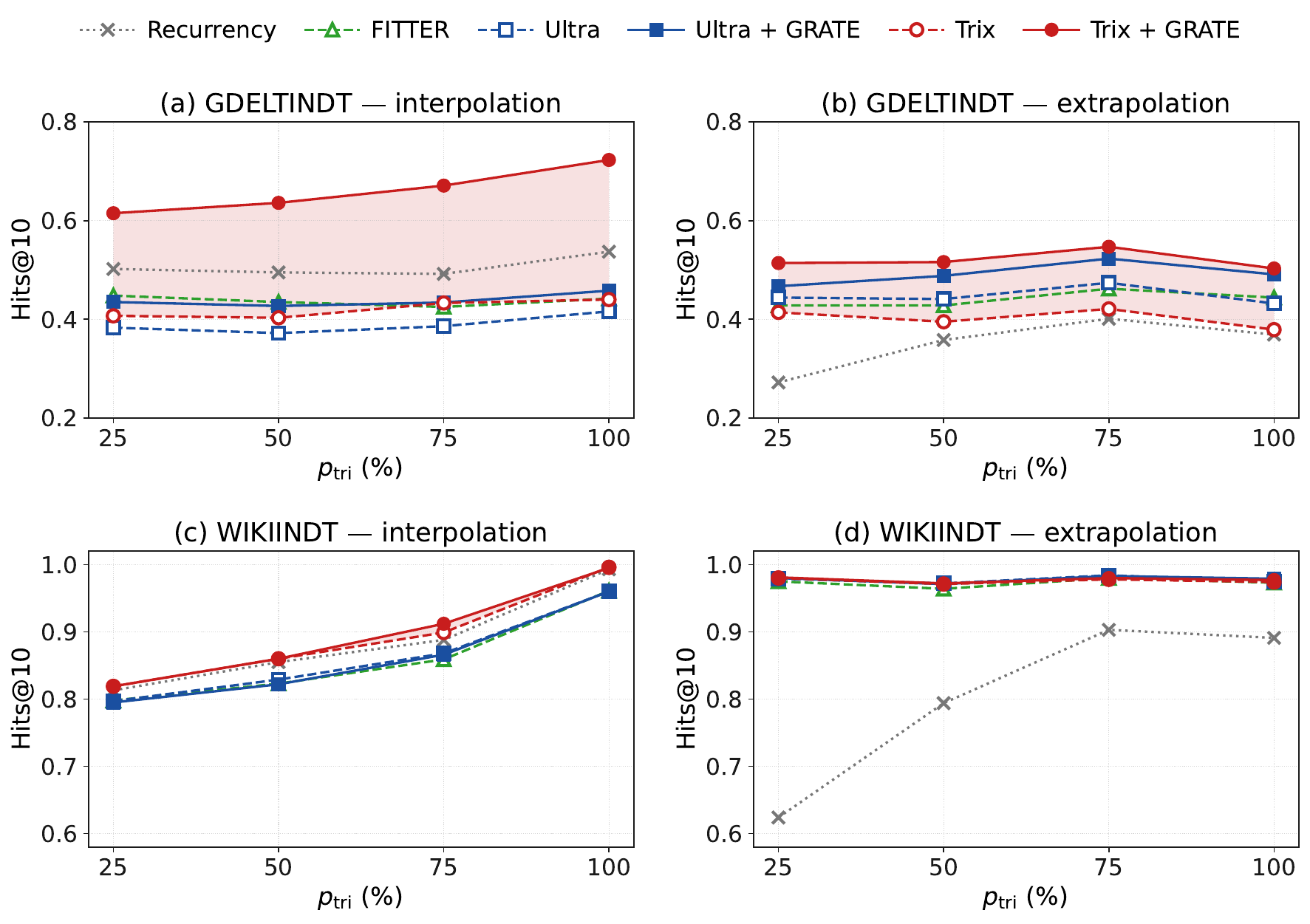}
  \caption{Hits@$10$ counterpart of \Cref{fig:sweep_results}.
    On \textsc{WIKIIndT}-interpolation the curves saturate near the
    ceiling, compressing the differences visible under MRR.}
  \label{fig:sweep_results_h10}
\end{figure*}

\Cref{fig:sweep_results_h1,fig:sweep_results_h10} plot Hits@$1$ and
Hits@$10$ in the same format as the MRR-based \Cref{fig:sweep_results}.

%% file: sections/app_implementation.tex
\section{Implementation and Base Model Details}
\label{app:implementation}

Unless explicitly noted otherwise, we follow the original
\textsc{Ultra} \citep{galkintowards} and \textsc{Trix}
\citep{zhangtrix} configurations and training recipes.

\subsection{Training Configuration}

\topic{Frequency basis.}
\Method{} uses the standard rotary frequency vector
$\omega_k = 1/10000^{2k/d}$ with $k = 0, 1, \dots, d/2 - 1$.
We keep the RoPE base at $10000$ in all experiments and do
not tune it.

\topic{Relative time scale.}
We pass the integer relative time gap
\[
  \Delta t_{uv} = \tau - t_{uv}
\]
directly to the RoPE rotation, without normalising it by
the dataset time range. This choice follows the design goal
of keeping \Method{} dataset-agnostic. RoPE already
contains frequencies spanning several orders of magnitude,
so low-frequency components can respond to large temporal
gaps while high-frequency components capture finer
differences. Moreover, normalising by a dataset-specific
time range would require access to target-graph statistics
at inference time. In our inductive transfer setting,
target timestamps are disjoint from training and may use
different granularities, such as daily timestamps in ICEWS
and yearly timestamps in WIKI. We therefore avoid
dataset-level rescaling and keep the temporal
transformation dependent only on relative time gaps.

\topic{Base models.}
\textsc{Ultra} and \textsc{Trix} use $L = 6$
message-passing layers, PNA aggregation, and hidden
dimension $d = 32$ \citep{galkintowards}. \Method{}
operates entirely inside the per-edge message function
and requires no changes to the aggregation or
update steps of the base model.

\subsection{Base Model Integration}

\Method{} integrates into any NBFNet-style base model as a
replacement for the per-edge message function $\phi$. We
instantiate on two:
\begin{itemize}
  \tightlistsetup
  \item \textbf{\textsc{Ultra}+\Method{}} replaces ULTRA's
    \citep{galkintowards} DistMult entity message with
    $m^{\Method{}}_{uv}$ from \Cref{eq:grate-message}.
    ULTRA's count-aggregated relation graph is unchanged.
  \item \textbf{\textsc{Trix}+\Method{}} performs the same
    swap on the TRIX base model \citep{zhangtrix}, which
    uses a role-aware relation graph that distinguishes
    head- and tail-side incidences.
\end{itemize}
\Method{} performs no surgery on the relation graph; it
is purely an entity-side modification. Whatever relation
features the base model produces, whether single-pass
like \textsc{Ultra} or iterative like \textsc{Trix}, are
consumed by the \Method{}-replaced message function
without modification.

\subsection{Hardware and Reproducibility}

All joint pretraining runs on ICEWS14\,+\,ICEWS05-15 use
$2 \times$ NVIDIA A100-80G GPUs. For \textsc{Ultra} and
\textsc{Ultra}\,+\,\Method{} we use batch size~8; for
\textsc{Trix} and \textsc{Trix}\,+\,\Method{} we use
batch size~2. Detailed per-epoch runtimes, memory usage,
and wall-clock time to peak MRR are reported in
\Cref{tab:efficiency}. Code and the constructed
\textsc{GDELTIndT} and \textsc{WIKIIndT} datasets are
available at
\url{https://anonymous.4open.science/r/GRATE-F431/README.md}.

%% file: sections/app_datasets.tex
\section{Dataset Construction and Statistics}
\label{app:datasets}

\subsection{Source Dataset Statistics}

\Cref{tab:source_datasets} lists statistics for the six
TKGs used in this paper. The inductive transfer splits we
construct from GDELT and WIKI are reported separately in
\Cref{tab:gdelt_sweep,tab:wiki_sweep}.

\begin{table*}[!tp]
  \centering
  \small
  \setlength{\tabcolsep}{6pt}
  \begin{tabular}{lcccccc}
    \toprule
    Dataset      & ICEWS14   & ICEWS05-15 & GDELT          & ICEWS18    & YAGO       & WIKI       \\
    \midrule
    Entities     &  7{,}128  &  10{,}488  &           500  &  23{,}033  &  10{,}623  &  12{,}554  \\
    Relations    &      230  &       251  &            20  &       256  &        10  &        24  \\
    Times        &      365  &  4{,}017   &           366  &       303  &       189  &       232  \\
    Train        & 72{,}826  & 386{,}962  & 2{,}735{,}685  & 373{,}018  & 161{,}540  & 539{,}286  \\
    Validation   &  8{,}941  &  46{,}275  &    341{,}961  &  45{,}995  &  19{,}523  &  67{,}538  \\
    Test         &  8{,}963  &  46{,}092  &    341{,}961  &  49{,}995  &  20{,}026  &  63{,}110  \\
    Granularity  & Daily     & Daily      & 15 min        & Daily      & Yearly     & Yearly     \\
    \bottomrule
  \end{tabular}
  \caption{Source dataset statistics for the six TKGs used
    in this paper.}
  \label{tab:source_datasets}
\end{table*}

\subsection{Split Protocol}
\label{app:split_protocol}

The model is trained on $\Gtrain$ and hyperparameters are
selected on a held-out $\Qvalid \subset \Qtrain$, keeping
the inference graph fully
unseen.\footnote{Two protocols exist: holding the
validation set out from the inference graph
\citep{lee2023ingram} or from the training graph
\citep{zhou2023multi}. We adopt the latter to eliminate
leakage from the target.} At inference, an observed subset
$\Qobs \subset \Qinf$ serves as the message-passing graph
and predictions are scored on a disjoint
$\Qtest \subset \Qinf$ with
$\Qobs \cap \Qtest = \emptyset$.

\subsection{Inductive Transfer Construction Procedure}

Starting from a source TKG $(V, R, \Tau, Q)$, we extend
INGRAM Algorithm~2 \citep{lee2023ingram} to quadruples
and add a chronological time split:
\begin{enumerate}
  \tightlistsetup
  \item Restrict to the giant connected component on the
    entity-level graph.
  \item Partition $R = R_{\mathrm{tr}} \cup R_{\mathrm{inf}}$
    disjointly with $|R_{\mathrm{inf}}| / |R| = p_R$.
  \item Partition timestamps \emph{chronologically}:
    $T_{\mathrm{tr}}$ contains the earliest
    $(1 - p_T) \cdot |\Tau|$ timestamps,
    $T_{\mathrm{inf}}$ the remaining $p_T \cdot |\Tau|$.
  \item Sample $n_{\mathrm{tr}}$ seed entities and
    BFS-expand with a per-hop neighbour cap to obtain a
    candidate $V_{\mathrm{tr}}$. The training edge set is
    $E_{\mathrm{tr}} = \{(s, r, o, \tau) : s, o \in
    V_{\mathrm{tr}}, r \in R_{\mathrm{tr}}, \tau \in
    T_{\mathrm{tr}}\}$; we take its giant component and
    define $V_{\mathrm{tr}}$ as the entities appearing in
    it.
  \item Remove $V_{\mathrm{tr}}$ from the source graph;
    sample $n_{\mathrm{inf}}$ seeds from the giant component
    of the remainder; expand to a candidate
    $V_{\mathrm{inf}}$.
  \item Form
    $E_{\mathrm{inf}}^{\mathrm{pool}} = \{(s, r, o, \tau) :
    s, o \in V_{\mathrm{inf}}\}$ and partition it into
    $X = \{r \in R_{\mathrm{tr}}, \tau \in T_{\mathrm{tr}}\}$
    (seen relation \emph{and} seen time) and
    $Y = \{r \in R_{\mathrm{inf}}, \tau \in T_{\mathrm{inf}}\}$
    (strictly disjoint). Subsample to obtain
    $|X|:|Y| = (1 - p_{\mathrm{tri}}) : p_{\mathrm{tri}}$.
    At $p_{\mathrm{tri}} = 1$ the inference graph is fully
    V/R/T-disjoint; smaller values mix in seen edges as
    additional supporting context.
  \item Split $E_{\mathrm{tr}}$ into \textsf{train.txt} /
    \textsf{valid.txt} (90/10) and split $E_{\mathrm{inf}}$
    into \textsf{msg.txt} / \textsf{test.txt} (80/20). The
    \textbf{inter} mode uses a random shuffle (msg and test
    span the same time window); the \textbf{extra} mode
    uses a chronological cut with
    $T_{\textsf{msg}} \le T_{\textsf{test}}$.
\end{enumerate}

\subsection{Per-source Sampling Parameters}

\Cref{tab:sampling} lists the per-source sampling
parameters. Both sources use $p_T = 0.3$ and split seed
$0$; $p_R = 0.5$ on GDELT and $p_R = 0.25$ on WIKI
(lowered because WIKI's $24$ relations would otherwise
leave too few for both $R_{\mathrm{tr}}$ and
$R_{\mathrm{inf}}$ post-GCC).

\begin{table}[t]
  \centering
  \footnotesize
  \setlength{\tabcolsep}{4pt}
  \begin{tabular}{lcccccc}
    \toprule
    Source & $p_R$ & $n_{\mathrm{tr}}$ & $n_{\mathrm{inf}}$ &
      hop\_cap & hops$_{\mathrm{tr}}$ & hops$_{\mathrm{inf}}$ \\
    \midrule
    GDELT & 0.50 &   50 &  300 &  5 & 1 & 1 \\
    WIKI  & 0.25 & 2000 & 1000 & 20 & 1 & 2 \\
    \bottomrule
  \end{tabular}
  \caption{Per-source sampling parameters used to generate
    the \textsc{GDELTIndT} and \textsc{WIKIIndT} benchmark
    suites.}
  \label{tab:sampling}
\end{table}

\topic{Temporal split boundaries.}
On GDELT, $T_{\mathrm{tr}}$ covers 2015-04-01 to
2015-12-12 ($256$ days) and $T_{\mathrm{inf}}$ covers
2015-12-13 to 2016-03-31 ($110$ days). On WIKI,
$T_{\mathrm{tr}}$ covers the first $162$ years and
$T_{\mathrm{inf}}$ the remaining $70$ years. The split
is by date or year (not by quadruple count), so the
inference timestamps are strictly later than every
training timestamp.

\subsection{\textsc{GDELTIndT} Statistics}

\Cref{tab:gdelt_sweep} reports per-variant entity, edge,
and split sizes for the eight \textsc{GDELTIndT}
variants. All eight share the V/R/T-disjoint partition;
only the inference-edge mix and the msg/test split differ.

\begin{table*}[!tp]
  \centering
  \footnotesize
  \setlength{\tabcolsep}{6pt}
  \begin{tabular}{lccrrrrrr}
    \toprule
    Variant & $p_{\mathrm{tri}}$ & mode & $|V_{\mathrm{tr}}|$ & $|V_{\mathrm{inf}}|$ & train & valid & msg & test \\
    \midrule
    -25-inter  & 0.25 & inter & 226 & 274 & 250.8 & 27.9 & 272.6 & 68.1 \\
    -50-inter  & 0.50 & inter & 226 & 274 & 209.9 & 23.3 & 268.4 & 67.1 \\
    -75-inter  & 0.75 & inter & 235 & 265 & 245.7 & 27.3 & 152.3 & 38.1 \\
    -100-inter & 1.00 & inter & 230 & 270 & 220.2 & 24.5 & 123.8 & 30.9 \\
    -25-extra  & 0.25 & extra & 225 & 275 & 209.4 & 23.3 & 318.4 & 79.6 \\
    -50-extra  & 0.50 & extra & 235 & 265 & 235.1 & 26.1 & 212.6 & 53.2 \\
    -75-extra  & 0.75 & extra & 227 & 273 & 237.9 & 26.4 & 172.0 & 43.0 \\
    -100-extra & 1.00 & extra & 233 & 267 & 239.2 & 26.6 & 114.6 & 28.6 \\
    \bottomrule
  \end{tabular}
  \caption{\textsc{GDELTIndT} benchmark suite ($p_R = 0.5$,
    $p_T = 0.3$). Edge counts in thousands.}
  \label{tab:gdelt_sweep}
\end{table*}

\subsection{\textsc{WIKIIndT} Statistics}

\Cref{tab:wiki_sweep} reports per-variant statistics for
the eight \textsc{WIKIIndT} variants. The drop in
$|V_{\mathrm{inf}}|$ at $p_{\mathrm{tri}} = 1$ reflects
that strict V/R/T disjointness on WIKI's small
$24$-relation vocabulary leaves a smaller candidate
inference set after the chronological time split.

\begin{table*}[!tp]
  \centering
  \footnotesize
  \setlength{\tabcolsep}{6pt}
  \begin{tabular}{lccrrrrrr}
    \toprule
    Variant & $p_{\mathrm{tri}}$ & mode & $|V_{\mathrm{tr}}|$ & $|V_{\mathrm{inf}}|$ & train & valid & msg & test \\
    \midrule
    -25-inter  & 0.25 & inter & 722 & 1{,}089 & 27.7 & 3.1 & 20.3 & 5.1 \\
    -50-inter  & 0.50 & inter & 765 &   942 & 28.2 & 3.1 &  9.9 & 2.5 \\
    -75-inter  & 0.75 & inter & 743 &   821 & 30.6 & 3.4 &  5.2 & 1.3 \\
    -100-inter & 1.00 & inter & 784 &   235 & 35.2 & 3.9 &  4.7 & 1.2 \\
    -25-extra  & 0.25 & extra & 873 & 1{,}040 & 34.6 & 3.8 & 19.4 & 4.8 \\
    -50-extra  & 0.50 & extra & 851 &   909 & 37.2 & 4.1 &  7.6 & 1.9 \\
    -75-extra  & 0.75 & extra & 768 &   843 & 33.4 & 3.7 &  6.3 & 1.6 \\
    -100-extra & 1.00 & extra & 855 &   240 & 35.8 & 4.0 &  4.8 & 1.2 \\
    \bottomrule
  \end{tabular}
  \caption{\textsc{WIKIIndT} benchmark suite
    ($p_R = 0.25$, $p_T = 0.3$). Edge counts in thousands.}
  \label{tab:wiki_sweep}
\end{table*}

\subsection{Benchmark Representativeness Analysis}
\label{app:representativeness}

To verify that our V/R/T-disjoint partitioning preserves
the structural properties of the source TKGs, we compare
degree distributions, temporal gap statistics, and triple
recurrence rates between the full source datasets and the
constructed \textsc{GDELTIndT} and \textsc{WIKIIndT}
benchmark variants.

\subsubsection{Full Source TKG Statistics}

\Cref{tab:source_structural} reports structural statistics
for the full GDELT and WIKI TKGs from which our inductive
benchmarks are derived.

\begin{table}[t]
  \centering
  \footnotesize
  \setlength{\tabcolsep}{4pt}
  \begin{tabular}{lcc}
    \toprule
    Statistic & GDELT & WIKI \\
    \midrule
    Entities           & 500        & 12{,}554 \\
    Relations          & 20         & 24 \\
    Timestamps         & 366        & 232 \\
    Total edges        & 3{,}419{,}607 & 669{,}934 \\
    Mean degree        & 13{,}678   & 108 \\
    Median degree      & 10{,}351   & 20 \\
    p95/median ratio   & 3.3        & 14.4 \\
    \midrule
    Recurring triples (\%) & 54.2   & 65.9 \\
    Avg occ./triple    & 3.7        & 20.0 \\
    \midrule
    \bottomrule
  \end{tabular}
  \caption{Structural statistics of the full GDELT and
    WIKI source TKGs. Degree is defined as the total
    number of edges incident to an entity across all
    timestamps. Recurring triples are unique $(s, r, o)$
    triples appearing at more than one timestamp.}
  \label{tab:source_structural}
\end{table}

\subsubsection{Comparison with IndT Variants}

\Cref{tab:representativeness} compares key structural
properties of the \textsc{GDELTIndT} and \textsc{WIKIIndT}
variants against their source TKGs. We report statistics
for the inference graph of each variant, as this is the
graph used for zero-shot evaluation.

\begin{table*}[!tp]
  \centering
  \footnotesize
  \setlength{\tabcolsep}{5pt}
  \begin{tabular}{llccccc}
    \toprule
    Source & Variant & Mean deg. & Med. deg. & p95/med. &
      Recur. (\%) & Avg occ. \\
    \midrule
    \multirow{5}{*}{GDELT}
    & Full TKG         & 13{,}678 & 10{,}351 & 3.3 & 54.2 & 3.7 \\
    \cmidrule(l){2-7}
    & IndT-25  & 1{,}990  & 1{,}498  & 3.3 & 42   & 2.4 \\
    & IndT-50  & 1{,}959  & 1{,}491  & 3.3 & 41   & 2.3 \\
    & IndT-75  & 1{,}149  & 881      & 3.3 & 35   & 2.1 \\
    & IndT-100 & 917      & 693      & 3.4 & 42   & 2.4 \\
    \midrule
    \multirow{5}{*}{WIKI}
    & Full TKG          & 108  & 20  & 14.4 & 65.9 & 20.0 \\
    \cmidrule(l){2-7}
    & IndT-25  & 38   & 23  & 3.6  & 89   & 17.2 \\
    & IndT-50  & 22   & 9   & 7.2  & 88   & 9.8  \\
    & IndT-75  & 13   & 3   & 19.7 & 65   & 6.6  \\
    & IndT-100 & 41   & 22  & 3.5  & 79   & 22.2 \\
    \bottomrule
  \end{tabular}
  \caption{Structural comparison between the full source
    TKGs and the inference graphs of the IndT benchmark
    variants (inter mode shown; extra mode yields similar
    statistics). The IndT variants have lower absolute
    degree due to relation partitioning ($p_R = 0.5$ for
    GDELT, $p_R = 0.25$ for WIKI) and time partitioning
    ($p_T = 0.3$), but preserve the distributional shape:
    GDELT variants retain a similar p95/median ratio,
    and recurrence patterns remain comparable to the
    source TKGs.}
  \label{tab:representativeness}
\end{table*}

\topic{Degree distribution.}
The full GDELT TKG has mean degree 13{,}678 and median
10{,}351, reflecting that nearly all 500 entities
participate in events at almost every timestamp.
\textsc{GDELTIndT} variants have lower absolute degree
(mean 917 to 1{,}990) because the relation partition
($p_R = 0.5$) and time partition ($p_T = 0.3$) retain only
edges matching the assigned subsets, but they preserve the
distributional shape: p95/median ratios are 3.3 to 3.4,
matching the full TKG. The full WIKI TKG has mean
degree 108 and median 20, reflecting a long-tailed
distribution typical of encyclopaedic knowledge. The
\textsc{WIKIIndT} variants (mean 13 to 41, median 3 to 23)
retain a long-tailed shape with varying skewness across
$p_{\mathrm{tri}}$ levels.

\topic{Recurrence.}
In the full GDELT TKG, 54.2\% of unique $(s, r, o)$
triples recur across multiple timestamps, with an average
of 3.7 occurrences per triple. \textsc{GDELTIndT} variants
show 35\% to 42\% recurrence (average 2.1 to 2.4
occurrences), lower due to relation and time
partitioning but preserving the same pattern. In the full
WIKI TKG, 65.9\% of triples recur with an average of 20.0
occurrences. \textsc{WIKIIndT} variants show 65\% to 89\%
recurrence (average 6.6 to 22.2 occurrences), comparable
to the source.

\topic{Summary.}
The primary design goal of these benchmarks is transfer
difficulty: V/R/T disjointness ensures no entity, relation,
or timestamp is shared between training and inference.
The analysis above confirms that despite the partitioning,
the resulting splits preserve the degree distributions
and recurrence patterns of their source TKG benchmarks.

%% file: sections/app_complexity.tex
\section{Parameter and Complexity Analysis}
\label{app:params}

\subsection{Theoretical Analysis}

\topic{Parameter count.}
\Method{} adds no learnable parameters over the base model.
The rotation $R(\Delta t)$ in \Cref{eq:rotated-message} is
parameterised entirely by the fixed RoPE frequency vector
$\omega_k = 1/10000^{2k/d}$; the gate $g_{uv}$ in
\Cref{eq:gate} is a parameter-free composition of a dot
product between $Q_q$ (which reuses the layer-$\ell{-}1$
head state and the existing query relation embedding) and
the rotated message, followed by a $1/\sqrt{d}$ scaling
and a sigmoid. \Cref{tab:params} reports parameter counts
on the two transductive ICEWS sources at the dimension
$d = 32$ used in the main paper. TKGE methods learn
per-entity, per-relation, and per-timestamp embeddings,
so their counts grow with $|V|$, $|R|$, and $|\Tau|$ and
vary substantially across datasets. \textsc{Ultra} and
\textsc{Trix} are vocabulary-agnostic foundation models
whose counts at fixed $(L, d)$ are dataset-independent,
and \Method{} adds zero on top of either.

\begin{table}[t]
  \centering
  \scriptsize
  \setlength{\tabcolsep}{4pt}
  \begin{tabular}{lcc}
    \toprule
    Model & ICEWS14 & ICEWS05-15 \\
    \midrule
    TComplEx   & $2.6 \times 10^6$ & $3.8 \times 10^6$ \\
    TNTComplEx & $2.7 \times 10^6$ & $3.9 \times 10^6$ \\
    \midrule
    \textsc{Ultra} \citep{galkintowards} & \multicolumn{2}{c}{$1.7 \times 10^5$ (dataset-independent)} \\
    $+$\,\Method{}                       & \multicolumn{2}{c}{$1.7 \times 10^5$ (\Method{} adds $0$)} \\
    \textsc{Trix} \citep{zhangtrix}      & \multicolumn{2}{c}{$3.4 \times 10^5$ (dataset-independent)} \\
    $+$\,\Method{}                       & \multicolumn{2}{c}{$3.4 \times 10^5$ (\Method{} adds $0$)} \\
    \bottomrule
  \end{tabular}
  \caption{Trainable parameter counts on the two
    transductive ICEWS sources. TKGE baselines
    \citep{tcomplexlacroix2020tensor} are dataset-specific.
    Base model counts at $L = 6$, $d = 32$, PNA aggregation
    are dataset-independent.}
  \label{tab:params}
\end{table}

\Method{}-augmented base models use roughly an order of
magnitude fewer parameters than the TKGE baselines on
ICEWS14 ($1.7$--$3.4 \times 10^5$ vs.\
$\sim$$2.7 \times 10^6$) and ICEWS05-15
($1.7$--$3.4 \times 10^5$ vs.\ $\sim$$3.9 \times 10^6$).
A parameter-free mechanism in time also has a structural
advantage under inductive transfer evaluation: there is no
parameter to mismatch when
$\Tinf \cap \Ttrain = \emptyset$, whereas TKGE methods
maintain a per-timestamp vocabulary of size $|\Ttrain|$
that is by definition unusable when the inference
timestamps are disjoint.

\topic{Time complexity.}
A static NBFNet-style base model performs $\mathcal{O}(Md)$
work per layer for the entity-side message and aggregation
\citep{zhu2021neural,galkintowards}. \Method{} adds, per
edge:
\begin{itemize}
  \tightlistsetup
  \item $\mathcal{O}(1)$ to compute
    $\Delta t_{uv} = \tau - t_{uv}$;
  \item $\mathcal{O}(d)$ to apply the pair-wise rotation
    $R(\Delta t_{uv}) m$ (\Cref{eq:rope-pair});
  \item $\mathcal{O}(d)$ for the dot product
    $Q_q^{\top} m^{\mathrm{rot}}_{uv}$ in the gate;
  \item $\mathcal{O}(1)$ for the sigmoid and the gate
    multiply.
\end{itemize}
Summing over edges, \Method{} contributes
$\mathcal{O}(Md)$ per layer (the same order as the host
base model), so the total per-query forward time is
$\mathcal{O}(LMd)$, matching the static base model
asymptotically.

\topic{Space complexity.}
\Method{} carries one persistent float per edge for $t_e$,
contributing $\mathcal{O}(M)$ persistent storage. Per-layer
transient tensors are $\mathcal{O}(Md)$ for the cos/sin
rotation factors and the rotated message, and
$\mathcal{O}(M)$ for the per-edge gate scalars (the same
order as the base model's per-layer activation memory).

\subsection{Empirical Efficiency}

\Cref{tab:efficiency} reports empirical training efficiency
on joint ICEWS14\,+\,ICEWS05-15 pretraining. \Method{} does
not increase GPU memory usage in our implementation:
\textsc{Ultra} and \textsc{Ultra}\,+\,\Method{} both use
32\,GB, while \textsc{Trix} and
\textsc{Trix}\,+\,\Method{} both use 70\,GB. The
per-epoch runtime increases because the rotation and gate
introduce additional per-edge tensor operations. For
\textsc{Trix}, this overhead is largely offset by faster
convergence to the best validation checkpoint, yielding a
similar wall-clock time to peak performance (8\,h\,30\,min
vs.\ 7\,h\,47\,min). For \textsc{Ultra}, the temporal
extension reaches peak performance in 3\,h\,10\,min
vs.\ 2\,h\,05\,min for the static backbone, remaining
within a single-node training budget.

\begin{table}[!tp]
  \centering
  \small
  \setlength{\tabcolsep}{4pt}
  \begin{tabular}{lcccc}
    \toprule
    Model & Params & Mem & Ep.\,(min) & Wall to peak \\
    \midrule
    \textsc{Ultra} & $1.7\!\times\!10^5$ & 32\,GB & 59  & 2\,h\,05\,min \\
    $+$\,\Method{} & same                & 32\,GB & 85  & 3\,h\,10\,min \\
    \midrule
    \textsc{Trix} & $3.4\!\times\!10^5$ & 70\,GB & 75  & 7\,h\,47\,min \\
    $+$\,\Method{} & same               & 70\,GB & 107 & 8\,h\,30\,min \\
    \bottomrule
  \end{tabular}
  \caption{Efficiency comparison of static backbones vs.\
    \Method{}-augmented variants on joint
    ICEWS14\,+\,ICEWS05-15 pretraining.
    \emph{Wall to peak} is total training time to the epoch
    with highest MRR on the held-out validation set.}
  \label{tab:efficiency}
\end{table}

Overall, \Method{} preserves the parameter efficiency and
memory footprint of the base models, while adding moderate
per-edge computation. This supports the main design goal:
introducing temporal reasoning into inductive KG foundation
models without adding timestamp-specific parameters or
sacrificing transferability.

%% file: sections/app_heldout_transductive.tex
\section{Transductive Forecasting Comparison on ICEWS18 and YAGO}
\label{app:transductive_heldout}

\Cref{tab:heldout_transductive} places \Method{} in the
context of transductive forecasting baselines on ICEWS18
and YAGO. The two groups are not directly comparable:
fine-tuned methods are trained on each target dataset and
have full access to its entity, relation, and timestamp
vocabulary; \Method{} uses a single checkpoint pretrained
on ICEWS14\,+\,ICEWS05-15 and is evaluated without any
target-specific fine-tuning. The \%~fine-tuned rows report each
zero-shot variant's score as a percentage of the strongest
fine-tuned baseline in that column, quantifying how much
of the supervised ceiling is recovered at zero cost.

Despite the setting mismatch, \textsc{Trix}\,+\,\Method{}
recovers $97.8\%$ / $97.2\%$ / $99.2\%$ of the best
fine-tuned baseline in MRR / Hits@$1$ / Hits@$10$ on YAGO, and
$83.7\%$ / $79.9\%$ / $87.6\%$ on ICEWS18.
\textsc{Ultra}\,+\,\Method{} reaches $90.4\%$ MRR on YAGO
and $69.6\%$ on ICEWS18. The larger gap on ICEWS18 is
consistent with that dataset being harder to generalise
to zero-shot, as discussed in \Cref{sec:exp:heldout}.

\begin{table}[t]
  \centering
  \scriptsize
  \setlength{\tabcolsep}{3pt}
  \resizebox{\columnwidth}{!}{%
  \begin{tabular}{l ccc ccc}
    \toprule
    & \multicolumn{3}{c}{ICEWS18} & \multicolumn{3}{c}{YAGO} \\
    \cmidrule(lr){2-4} \cmidrule(lr){5-7}
    Method & MRR & H@1 & H@10 & MRR & H@1 & H@10 \\
    \midrule
    \multicolumn{7}{l}{\emph{Fine-tuned on target}} \\
    RE-GCN       & \textbf{.326} & \textbf{.224} & \textbf{.526} & .822 & .787 & .885 \\
    xERTE        & .292 & .209 & .463 & .873 & .842 & .912 \\
    TLogic       & .296 & .204 & .481 & .765 & .740 & .792 \\
    TANGO        & .284 & .191 & .463 & .624 & .590 & .678 \\
    TimeTraveler & .291 & .213 & .439 & \textbf{.877} & \textbf{.846} & .912 \\
    \midrule
    \multicolumn{7}{l}{\emph{Zero-shot (single pretrained checkpoint)}} \\
    FITTER                   & .228 & .128 & .444 & .792 & .736 & .886 \\
    \midrule
    \textsc{Ultra}           & .180 & .092 & .369 & .726 & .647 & .872 \\
    $+$\,\Method{}           & .227 & .126 & .447 & .793 & .738 & .890 \\
    \%~fine-tuned (\textsc{Ultra}) & $69.6\%$ & $56.3\%$ & $85.0\%$ & $90.4\%$ & $87.2\%$ & $97.6\%$ \\
    \midrule
    \textsc{Trix}            & .274 & .170 & .484 & .800 & .740 & \textbf{.916} \\
    $+$\,\Method{}           & .273 & .179 & .461 & .858 & .822 & .905 \\
    \%~fine-tuned (\textsc{Trix})  & $83.7\%$ & $79.9\%$ & $87.6\%$ & $97.8\%$ & $97.2\%$ & $99.2\%$ \\
    \bottomrule
  \end{tabular}}
  \caption{Zero-shot transfer to ICEWS18 and YAGO vs.\
    transductive forecasting baselines from
    \citet{gastinger2023comparing} under the
    single-step extrapolation protocol.
    \emph{Fine-tuned} methods are trained directly on each
    target; \emph{zero-shot} methods use a single pretrained
    checkpoint with no fine-tuning.
    FITTER uses the ICEWS14 checkpoint from the original paper.
    \%~fine-tuned rows report
    the zero-shot score as a percentage of the best
    fine-tuned baseline per column.
    Best per column \textbf{bold}.}
  \label{tab:heldout_transductive}
\end{table}

%% file: sections/app_transductive_full.tex
\section{Full Transductive Comparison}
\label{app:transductive_full}

\Cref{tab:transductive_full} extends \Cref{tab:transductive} with
dataset-specific transductive TKG embedding baselines trained and
evaluated on each dataset independently. These methods learn
per-entity, per-relation, and per-timestamp embeddings and are not
transferable across datasets; they serve as an in-distribution fine-tuned
reference.

\begin{table}[t]
  \centering
  \scriptsize
  \setlength{\tabcolsep}{1pt}
  \begin{tabular}{l ccc ccc}
    \toprule
    & \multicolumn{3}{c}{ICEWS14} & \multicolumn{3}{c}{ICEWS05-15} \\
    \cmidrule(lr){2-4} \cmidrule(lr){5-7}
    Method & MRR & H@1 & H@10 & MRR & H@1 & H@10 \\
    \midrule
    TA-DistMult     & .477 & .363 & .686 & .474 & .346 & .728 \\
    TeRo            & .562 & .468 & .732 & .586 & .469 & .795 \\
    TComplEx        & .610 & \textbf{.530} & .770 & .660 & \textbf{.590} & .800 \\
    TNTComplEx      & \textbf{.620} & .520 & .760 & \textbf{.670} & \textbf{.590} & .810 \\
    \midrule
    FITTER              & .583 & ---   & .760 & .590 & ---   & .787 \\
    \midrule
    \textsc{Ultra}              & .495 & .373 & .731 & .451 & .316 & .720 \\
    $+$\,\Method{} & \textbf{.620} & .520 & \textbf{.793} & .621 & .500 & .819 \\
    $\%\,\Delta$                     & $+25.3\%$ & $+39.4\%$ & $+8.5\%$ & $+37.7\%$ & $+58.2\%$ & $+13.8\%$ \\
    \midrule
    \textsc{Trix}               & .486 & .366 & .719 & .456 & .322 & .722 \\
    $+$\,\Method{} & .609 & .507 & \textbf{.793} & .617 & .501 & \textbf{.830} \\
    $\%\,\Delta$                     & $+25.3\%$ & $+38.5\%$ & $+10.3\%$ & $+35.3\%$ & $+55.6\%$ & $+15.0\%$ \\
    \bottomrule
  \end{tabular}
  \caption{Transductive link prediction on ICEWS14 and ICEWS05-15,
    including dataset-specific TKG embedding baselines.
    FITTER is trained and tuned separately on each dataset;
    \textsc{Ultra} and \textsc{Trix} are retrained on the joint
    ICEWS14\,+\,ICEWS05-15 pretraining mix without the \Method{}
    module. $\%\,\Delta$ rows show the relative improvement of
    \Method{} over the corresponding base model.
    Best per column \textbf{bold}.}
  \label{tab:transductive_full}
\end{table}

\topic{Comparison with TA-DistMult.}
TA-DistMult also uses a DistMult-style message function but
incorporates time as a static additive component in the scoring
function, which cannot model the varying relevance of facts at
different time offsets. \Method{} improves substantially over
TA-DistMult on both datasets, showing that encoding temporal
displacement via relative-time rotation provides a stronger
temporal signal than additive time features alone.

\topic{Comparison with TeRo.}
TeRo also injects a rotation into the temporal scoring function,
but applies a per-timestamp rotation that must be learned during
training, making it inapplicable to unseen timestamps.
\Method{} instead rotates by the relative time gap using a fixed
frequency basis (RoPE), requiring no timestamp-specific parameters
and generalising to arbitrary timestamps at inference.
\Method{} outperforms TeRo on both datasets despite this
additional constraint.

\topic{Comparison with TComplEx and TNTComplEx.}
TComplEx and TNTComplEx are the strongest transductive baselines,
relying on per-entity, per-relation, and per-timestamp embeddings
that scale with vocabulary size (on the order of $10^6$
parameters). \Method{} adds zero learnable parameters over its
base model (\textsc{Ultra}\,+\,\Method{}: $1.7{\times}10^5$
parameters at dim~32; \textsc{Trix}\,+\,\Method{}:
$3.4{\times}10^5$), yet matches or comes within a few points of
their performance on both datasets, demonstrating that temporal
signal can be captured through relative-time alignment without
dataset-specific embedding tables.

%% file: sections/app_baseline_inapplicability.tex
\section{Baseline Inapplicability Analysis}
\label{app:baseline_inapplicability}

Several classes of temporal models might appear to be
natural baselines for the V/R/T-disjoint inductive
transfer setting. We explain below why each class is
architecturally incompatible with this evaluation
protocol.

\subsection{Transductive TKG Models}
\label{app:baseline:transductive_tkg}

Both interpolation and extrapolation TKG models learn
entity-indexed or relation-indexed parameters that are
integral to their architectures.

Interpolation models (TKG completion) typically extend
static KG scoring functions with temporal components:

\begin{itemize}
  \tightlistsetup
  \item \textbf{TTransE} \citep{leblay2018deriving} and
    \textbf{TA-DistMult}
    \citep{garcia-duran-etal-2018-learning}: learn
    per-entity and per-relation embeddings as part of
    the scoring function.
  \item \textbf{TComplEx} and \textbf{TNTComplEx}
    \citep{tcomplexlacroix2020tensor}: factorise a
    four-dimensional tensor indexed by the entity and
    relation vocabulary.
\end{itemize}

Extrapolation models (TKG forecasting) similarly rely
on entity-indexed components. We verified via source
code that this holds for representative models in this
category:

\begin{itemize}
  \tightlistsetup
  \item \textbf{RE-NET} \citep{jin2020recurrent}: maintains
    per-entity recurrent states (\texttt{ent\_embeds})
    that track each entity's event history, undefined
    for entities not seen during training.
  \item \textbf{CyGNet} \citep{zhu2021learning}: uses a
    per-entity co-occurrence matrix for its copy-mode
    generation mechanism, requiring entity-indexed
    statistics from the training graph.
  \item \textbf{TiRGN} \citep{li2022tirgn}: maintains
    per-entity dynamic embeddings via GCN+GRU
    (\texttt{dynamic\_emb}), undefined at inference on a
    disjoint graph.
\end{itemize}

These entity-indexed parameters are not peripheral
components that can be removed without changing the
model. Replacing them would require substituting the
entire entity representation with a vocabulary-agnostic
alternative such as relation-graph conditioning (as in
ULTRA/TRIX), which would effectively reconstruct
our approach rather than constitute an independent
baseline.

\subsection{Temporal Graph Learning Models}
\label{app:baseline:temporal_graph}

Most temporal graph models are designed for dynamic
graph learning tasks (e.g., dynamic node classification,
future link prediction on evolving social or interaction
networks) rather than knowledge graph link prediction.
Representative models such as TGAT
\citep{xu2020inductive} and TGN
\citep{rossi2020temporal} operate on graphs with a fixed
or growing node set with numerical node features and do
not model \emph{typed relations}, a core component of KG
reasoning.

Adapting such models to the V/R/T-disjoint setting would
require addressing two barriers simultaneously. First,
their node-identity-dependent components (e.g., TGN's
per-node memory module, TGAT's per-node features) must
be replaced. Second, a mechanism for relation-typed
message passing must be introduced, which is outside
their design scope. Even if both issues were resolved,
the V/R/T-disjoint setting requires transfer to entirely
unseen entities, relations, and timestamps, whereas
these models assume at least partial overlap with
training entities.

\subsection{Other KG Foundation Models}
\label{app:baseline:other_kgfm}

Beyond ULTRA and TRIX, other KG foundation models
support transfer across unseen entities and relations.
\Method{} requires a clean per-edge message interface
(\Cref{eq:base-message}) into which the rotation and
gating can be inserted as a drop-in replacement. Two
recent models lack this interface:

\begin{itemize}
  \tightlistsetup
  \item \textbf{FLOCK} \citep{kim2026flock}: uses random
    walks combined with a sequence model rather than
    explicit per-edge message passing.
  \item \textbf{GraphOracle} \citep{du2026graphoracle}:
    uses a two-stage architecture in which multi-head
    attention over a Relation-Dependency Graph produces
    relation embeddings, followed by a separate
    entity-level GNN. Temporal information would need to
    be incorporated at multiple points rather than a
    single message function.
\end{itemize}

Extending \Method{} to architectures without explicit
per-edge messages is an interesting direction for future
work.

%% file: sections/app_theoretical.tex
\section{Theoretical Properties of the Rotary Encoding}
\label{app:theoretical}

In \Cref{sec:method} we introduced the \Method{} mechanism,
which rotates messages according to relative time gaps
and gates them by query alignment. We now formally establish
three properties of this construction that underpin its
effectiveness for temporal knowledge graph reasoning: norm
preservation under rotation, relative-time invariance of
inner products, and temporal discriminability of rotated
messages.

\subsection{Norm Preservation}
\label{app:theoretical:norm}

A desirable property of any temporal encoding is that it
should not distort the magnitude of the underlying
structural message. If temporal encoding amplified or
attenuated message norms, downstream aggregation would
conflate temporal influence with structural importance. The
following proposition shows that the rotary encoding in
\Cref{eq:rotated-message} satisfies this requirement
exactly.

\begin{proposition}[Norm Preservation]
\label{prop:norm-preservation}
Let $R(\delta) \in \mathbb{R}^{d \times d}$ be the
block-diagonal rotation matrix defined in
\Cref{eq:rope-matrix}, with $2 \times 2$ blocks
$R_k(\delta)$ as in \Cref{eq:rope-rotation}. For any
position offset $\delta \in \mathbb{R}$ and any vector
$m \in \mathbb{R}^d$,
\[
\|R(\delta)\, m\| = \|m\|.
\]
\end{proposition}

\begin{proof}
Since each block is a planar rotation matrix, it suffices
to verify that $R_k(\delta)$ is orthogonal, i.e.,
$R_k(\delta)^\top R_k(\delta) = I_2$. Writing
$\theta_k = \omega_k \delta$, we compute
\begin{align*}
&R_k(\delta)^\top R_k(\delta) \\
&= \begin{pmatrix} \cos\theta_k & \sin\theta_k \\
   -\sin\theta_k & \cos\theta_k \end{pmatrix}
   \begin{pmatrix} \cos\theta_k & -\sin\theta_k \\
   \sin\theta_k & \phantom{-}\cos\theta_k \end{pmatrix} \\
&= \begin{pmatrix} \cos^2\theta_k + \sin^2\theta_k & 0 \\
   0 & \sin^2\theta_k + \cos^2\theta_k \end{pmatrix}
 = I_2.
\end{align*}
Since $R(\delta) = \mathrm{diag}(R_0(\delta), \ldots,
R_{d/2-1}(\delta))$ is block-diagonal with orthogonal
blocks, $R(\delta)$ is itself orthogonal:
$R(\delta)^\top R(\delta) = I_d$. Therefore,
\begin{align*}
\|R(\delta)\, m\|^2
&= m^\top R(\delta)^\top R(\delta)\, m \\
&= m^\top m = \|m\|^2,
\end{align*}
and the result follows by taking square roots.
\end{proof}

\Cref{prop:norm-preservation} guarantees that the rotary
encoding in \Cref{eq:rotated-message} transforms the
direction of the message vector while leaving its magnitude
unchanged. In the context of \Method{}, this means that the
rotation preserves the message norm while reorienting the
structural representation $m_{uv} = h_u \odot r_{uv}$
according to the relative time gap. This property contrasts with element-wise
sinusoidal alternatives of the form
$m \odot \mathrm{SinCos}(\Delta t)$, which rescale each
dimension independently and can attenuate or amplify the
message norm depending on the time gap.

\subsection{Relative-Time Invariance}
\label{app:theoretical:invariance}

The central motivation for using rotary encoding in temporal
knowledge graphs is that temporal reasoning should depend on
relative time differences rather than absolute timestamps.
The following proposition formalises this property by showing
that the inner product between two rotated vectors depends
only on the difference of their position offsets.

\begin{proposition}[Relative-Time Invariance of Inner
  Products]
\label{prop:relative-invariance}
Let $R(\delta)$ be the rotation matrix from
\Cref{eq:rope-matrix}. For any vectors
$m, m' \in \mathbb{R}^d$ and any offsets
$\delta_1, \delta_2 \in \mathbb{R}$,
\[
\langle R(\delta_1)\, m,\; R(\delta_2)\, m' \rangle
= \langle R(\delta_1 - \delta_2)\, m,\; m' \rangle.
\]
In particular, the inner product depends only on the
relative offset $\delta_1 - \delta_2$, not on $\delta_1$
and $\delta_2$ individually.
\end{proposition}

\begin{proof}
Since each block $R_k(\delta)$ is a planar rotation, its
transpose satisfies $R_k(\delta)^\top = R_k(-\delta)$, and
the composition of two rotations obeys the group law
$R_k(\alpha)\, R_k(\beta) = R_k(\alpha + \beta)$.
Therefore,
\begin{align*}
&R(\delta_1)^\top R(\delta_2) \\
&= \mathrm{diag}\bigl(R_0(\delta_2 {-} \delta_1),\,
   \ldots,\,
   R_{d/2-1}(\delta_2 {-} \delta_1)\bigr) \\
&= R(\delta_2 - \delta_1).
\end{align*}
Applying this to the left-hand side:
\begin{align*}
&\langle R(\delta_1)\, m,\;
  R(\delta_2)\, m' \rangle \\
&= m^\top R(\delta_1)^\top R(\delta_2)\, m'
 = m^\top R(\delta_2 {-} \delta_1)\, m'.
\end{align*}
For the right-hand side, using
$R(\alpha)^\top = R(-\alpha)$:
\begin{align*}
&\langle R(\delta_1 {-} \delta_2)\, m,\;
  m' \rangle \\
&= m^\top R(\delta_1 {-} \delta_2)^\top m'
 = m^\top R(\delta_2 {-} \delta_1)\, m'.
\end{align*}
The two expressions are identical, which completes the
proof.
\end{proof}

\Cref{prop:relative-invariance} has a direct consequence for
\Method{}. The gate value in \Cref{eq:gate} computes
$Q_q^\top R(\Delta t_{uv})\, m_{uv}$, where
$\Delta t_{uv} = \tau - t_{uv}$. Since the gate depends on
timestamps only through the relative gap $\Delta t_{uv}$,
it is invariant under uniform shifts of all timestamps: if
both the query time $\tau$ and every edge timestamp $t_{uv}$
are shifted by the same constant, each relative gap remains
unchanged, and so does the gate value. This property eliminates dependence on absolute timestamps,
making the temporal encoding naturally applicable across
domains with different temporal origins.

\subsection{Temporal Discriminability}
\label{app:theoretical:discriminability}

Norm preservation and relative-time invariance establish
that the rotary encoding is well-behaved, but a temporal
encoding must also be \emph{expressive}: distinct time gaps
should produce distinct rotated messages. Without this
property, the model could not distinguish temporally close
supporting facts from distant ones. The following
proposition provides a sufficient condition under which
distinct time gaps yield distinct rotated messages.

\begin{proposition}[Temporal Discriminability]
\label{prop:temporal-discriminability}
Let $m \in \mathbb{R}^d$ be a non-zero vector and let
$\Delta t_1, \Delta t_2 \in \mathbb{R}$ with
$\Delta t_1 \neq \Delta t_2$. If there exists at least one
frequency index $k \in \{0, \ldots, d/2-1\}$ such that
\begin{equation}
    \omega_k\, (\Delta t_1 - \Delta t_2)
    \not\equiv 0 \pmod{2\pi}
    \label{eq:discriminability-condition}
\end{equation}
and the corresponding component pair
$(m_{2k},\, m_{2k+1}) \neq (0,\, 0)$, then
\[
R(\Delta t_1)\, m \neq R(\Delta t_2)\, m.
\]
\end{proposition}

\begin{proof}
Suppose for contradiction that
$R(\Delta t_1)\, m = R(\Delta t_2)\, m$. Then
$[R(\Delta t_1) - R(\Delta t_2)]\, m = 0$. Since $R$ is
block-diagonal, this reduces to
\[
[R_k(\Delta t_1) - R_k(\Delta t_2)]
\begin{pmatrix} m_{2k} \\ m_{2k+1} \end{pmatrix}
= \begin{pmatrix} 0 \\ 0 \end{pmatrix}
\]
for each frequency index $k$. Writing
$\theta_i = \omega_k\, \Delta t_i$ for $i = 1, 2$, the
difference of rotation blocks is
\begin{align*}
&R_k(\Delta t_1) - R_k(\Delta t_2) \\
&= \begin{pmatrix}
  \cos\theta_1 - \cos\theta_2 &
  -\sin\theta_1 + \sin\theta_2 \\
  \sin\theta_1 - \sin\theta_2 &
  \phantom{-}\cos\theta_1 - \cos\theta_2
\end{pmatrix}.
\end{align*}
The determinant of this matrix is
\begin{align}
    &\det[R_k(\Delta t_1) - R_k(\Delta t_2)] \nonumber \\
    &= (\cos\theta_1 - \cos\theta_2)^2
       + (\sin\theta_1 - \sin\theta_2)^2 \nonumber \\
    &= 2\,(1 - \cos(\theta_1 - \theta_2)).
    \label{eq:det-rotation-diff}
\end{align}
When $\theta_1 - \theta_2 =
\omega_k(\Delta t_1 - \Delta t_2) \not\equiv 0
\pmod{2\pi}$, we have $\cos(\theta_1 - \theta_2) \neq 1$,
so the determinant is strictly positive. A matrix with
positive determinant is invertible, and therefore its null
space is trivial: the only solution to
$[R_k(\Delta t_1) - R_k(\Delta t_2)]\, v = 0$ is $v = 0$.
Since $(m_{2k}, m_{2k+1}) \neq (0, 0)$ by hypothesis, we
obtain a contradiction.
\end{proof}

The condition in \Cref{eq:discriminability-condition} is
mild in practice. With the standard RoPE frequency set
$\omega_k = 10000^{-2k/d}$ from \Cref{eq:rope-freq}, the
period of the $k$-th frequency is
$2\pi / \omega_k = 2\pi \cdot 10000^{2k/d}$. For the
highest frequency ($k = 0$), the period is
$2\pi \approx 6.28$; for the lowest ($k = d/2 - 1$), it
approaches $2\pi \cdot 10^4$. A time gap difference
$|\Delta t_1 - \Delta t_2|$ would need to be an exact
integer multiple of every period in this exponentially
spaced set to violate the condition, which is a zero-measure
event for continuous-valued time gaps and unlikely to occur
in practice for integer-valued timestamps in any practical
TKG.

\medskip

Because the gate in \Cref{eq:gate} directly operates on
the rotated message, these properties govern how temporal
information influences message selection in \Method{}.
Together, the three propositions characterise the rotary
temporal encoding as a well-behaved and expressive
mechanism.
\Cref{prop:norm-preservation} ensures that temporal encoding
does not distort structural messages.
\Cref{prop:relative-invariance} guarantees that inner
products of rotated representations depend only on relative
time offsets, a property that underpins the shift-invariance
of the gating mechanism and enables cross-domain transfer.
\Cref{prop:temporal-discriminability} ensures that distinct
temporal contexts remain distinguishable under mild
conditions, allowing the model to leverage fine-grained
temporal information for prediction. These properties hold by construction and do not
depend on learned parameters, which is consistent with the
parameter-free design of \Method{}.

%% file: sections/app_beta_sensitivity.tex
\section{RoPE Base Frequency Sensitivity}
\label{app:beta}

\Method{} inherits from RoPE a single hyperparameter: the
base frequency $\beta$ that controls the frequency spectrum
$\omega_k = \beta^{-2k/d}$ (\Cref{eq:rope-freq}). We train
\textsc{Trix}\,+\,\Method{} with the standard LLM default
$\beta = 10\,000$ and evaluate on ICEWS14 by swapping $\beta$
at inference time without retraining, varying it across two
orders of magnitude from $1\,000$ to $100\,000$.

\begin{figure}[t]
  \centering
  \includegraphics[width=\columnwidth]{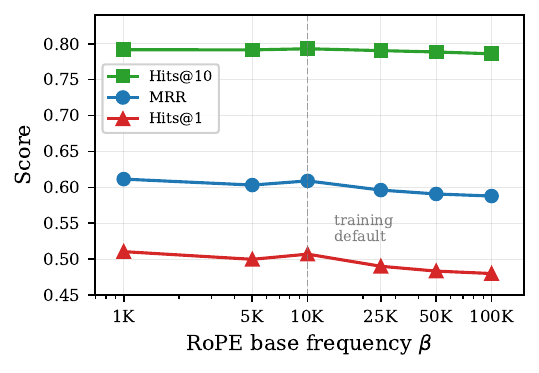}
  \caption{Sensitivity of \Method{} to the RoPE base
    frequency $\beta$ on ICEWS14. The model is trained once
    at $\beta = 10\,000$ (dashed line) and evaluated with
    different $\beta$ values at inference time. MRR varies by
    only $0.024$ across two orders of magnitude.}
  \label{fig:beta_sensitivity}
\end{figure}

\Cref{fig:beta_sensitivity} shows that all three metrics
vary within a narrow band: MRR spans $0.588$--$0.612$, a
total variation of $0.024$ across a $100\times$ range of
$\beta$, confirming that \Method{} is not sensitive to the
choice of base frequency.
The training default $\beta = 10\,000$ sits within $0.003$
MRR of the optimum at $\beta = 1\,000$, confirming that
\Method{} does not require careful tuning of this parameter.
This finding is consistent with the observation in FITTER
\citep{pan2026fittervocabularyagnosticcrossdomaininference},
where frequency-based temporal encodings similarly exhibit
low sensitivity to base frequency choices. The robustness
arises because the multi-frequency spectrum covers a broad
range of periodicities regardless of the exact base: shifting
$\beta$ rescales all frequencies uniformly, preserving the
relative spacing that drives temporal discrimination
(\Cref{prop:temporal-discriminability}).

%% file: references.bib
@article{su2024roformer,
  title={Roformer: Enhanced transformer with rotary position embedding},
  author={Su, Jianlin and Ahmed, Murtadha and Lu, Yu and Pan, Shengfeng and Bo, Wen and Liu, Yunfeng},
  journal={Neurocomputing},
  volume={568},
  pages={127063},
  year={2024},
  publisher={Elsevier}
}

@inproceedings{gastinger2024history,
  title={History Repeats Itself: A Baseline for Temporal Knowledge Graph Forecasting},
  author={Gastinger, Julia and Meilicke, Christian and Errica, Federico and Sztyler, Timo and Sch{\"u}lke, Anett and Stuckenschmidt, Heiner},
  booktitle={IJCAI},
  year={2024}
}

@inproceedings{du2026graphoracle,
  title={GraphOracle: Efficient Fully-Inductive Knowledge Graph Reasoning via Relation-Dependency Graphs},
  author={Du, Enjun and Liu, Siyi and Zhang, Yongqi},
  booktitle={Proceedings of the AAAI Conference on Artificial Intelligence},
  volume={40},
  number={23},
  pages={19055--19063},
  year={2026}
}

@inproceedings{zhou2023multi,
  title={A Multi-Task Perspective for Link Prediction with New Relation Types and Nodes},
  author={Zhou, Jincheng and Bevilacqua, Beatrice and Ribeiro, Bruno},
  booktitle={NeurIPS 2023 Workshop: New Frontiers in Graph Learning},
  year={2023}
}

@inproceedings{li2021temporal,
  title={Temporal knowledge graph reasoning based on evolutional representation learning},
  author={Li, Zixuan and Jin, Xiaolong and Li, Wei and Guan, Saiping and Guo, Jiafeng and Shen, Huawei and Wang, Yuanzhuo and Cheng, Xueqi},
  booktitle={Proceedings of the 44th international ACM SIGIR conference on research and development in information retrieval},
  pages={408--417},
  year={2021}
}

@inproceedings{
tcomplexlacroix2020tensor,
title={Tensor Decompositions for Temporal Knowledge Base Completion},
author={Timothée Lacroix and Guillaume Obozinski and Nicolas Usunier},
booktitle={International Conference on Learning Representations},
year={2020},
url={https://openreview.net/forum?id=rke2P1BFwS}
}

@inproceedings{jin2020recurrent,
  title={Recurrent Event Network: Autoregressive Structure Inference over Temporal Knowledge Graphs},
  author={Jin, Woojeong and Qu, Meng and Jin, Xisen and Ren, Xiang},
  booktitle={Proceedings of the 2020 Conference on Empirical Methods in Natural Language Processing (EMNLP)},
  pages={6669--6683},
  year={2020}
}

@inproceedings{tltcomplexzhang2022along,
  title={Along the Time: Timeline-traced Embedding for Temporal Knowledge Graph Completion},
  author={Zhang, Fuwei and Zhang, Zhao and Ao, Xiang and Zhuang, Fuzhen and Xu, Yongjun and He, Qing},
  booktitle={Proceedings of the 31st ACM International Conference on Information \& Knowledge Management},
  pages={2529--2538},
  year={2022}
}

@inproceedings{trivedi2017know,
  title={Know-evolve: Deep temporal reasoning for dynamic knowledge graphs},
  author={Trivedi, Rakshit and Dai, Hanjun and Wang, Yichen and Song, Le},
  booktitle={international conference on machine learning},
  pages={3462--3471},
  year={2017},
  organization={PMLR}
}

@inproceedings{xu2020tero,
  title={TeRo: A Time-aware Knowledge Graph Embedding via Temporal Rotation},
  author={Xu, Chengjin and Nayyeri, Mojtaba and Alkhoury, Fouad and Yazdi, Hamed Shariat and Lehmann, Jens},
  booktitle={Proceedings of the 28th International Conference on Computational Linguistics},
  pages={1583--1593},
  year={2020}
}

@inproceedings{goel2020diachronic,
  title={Diachronic embedding for temporal knowledge graph completion},
  author={Goel, Rishab and Kazemi, Seyed Mehran and Brubaker, Marcus and Poupart, Pascal},
  booktitle={Proceedings of the AAAI Conference on Artificial Intelligence},
  volume={34},

  pages={3988--3995},
  year={2020}
}

@inproceedings{leblay2018deriving,
  title={Deriving validity time in knowledge graph},
  author={Leblay, Julien and Chekol, Melisachew Wudage},
  booktitle={Companion proceedings of the the web conference 2018},
  pages={1771--1776},
  year={2018}
}

@article{lautenschlager2015icews,
  title={Icews event aggregations},
  author={Lautenschlager, Jennifer and Shellman, Steve and Ward, Michael},
  journal={Harvard Dataverse},
  volume={3},
  number={595},
  pages={28},
  year={2015}
}

@inproceedings{garcia-duran-etal-2018-learning,
    title = "Learning Sequence Encoders for Temporal Knowledge Graph Completion",
    author = "Garc{\'i}a-Dur{\'a}n, Alberto  and
      Duman{\v{c}}i{\'c}, Sebastijan  and
      Niepert, Mathias",
    editor = "Riloff, Ellen  and
      Chiang, David  and
      Hockenmaier, Julia  and
      Tsujii, Jun{'}ichi",
    booktitle = "Proceedings of the 2018 Conference on Empirical Methods in Natural Language Processing",
    month = oct # "-" # nov,
    year = "2018",
    address = "Brussels, Belgium",
    publisher = "Association for Computational Linguistics",
    url = "https://aclanthology.org/D18-1516/",
    doi = "10.18653/v1/D18-1516",
    pages = "4816--4821"
}

@inproceedings{galkintowards,
  title={Towards Foundation Models for Knowledge Graph Reasoning},
  author={Galkin, Mikhail and Yuan, Xinyu and Mostafa, Hesham and Tang, Jian and Zhu, Zhaocheng},
  booktitle={The Twelfth International Conference on Learning Representations},
 year={2024}
}

@inproceedings{liang2023learn,
  title={Learn from relational correlations and periodic events for temporal knowledge graph reasoning},
  author={Liang, Ke and Meng, Lingyuan and Liu, Meng and Liu, Yue and Tu, Wenxuan and Wang, Siwei and Zhou, Sihang and Liu, Xinwang},
  booktitle={Proceedings of the 46th international ACM SIGIR conference on research and development in information retrieval},
  pages={1559--1568},
  year={2023}
}

@inproceedings{sun2021timetraveler,
  title={TimeTraveler: Reinforcement Learning for Temporal Knowledge Graph Forecasting},
  author={Sun, Haohai and Zhong, Jialun and Ma, Yunpu and Han, Zhen and He, Kun},
  booktitle={Proceedings of the 2021 Conference on Empirical Methods in Natural Language Processing},
  pages={8306--8319},
  year={2021}
}

@inproceedings{liu2022tlogic,
  title={Tlogic: Temporal logical rules for explainable link forecasting on temporal knowledge graphs},
  author={Liu, Yushan and Ma, Yunpu and Hildebrandt, Marcel and Joblin, Mitchell and Tresp, Volker},
  booktitle={Proceedings of the AAAI conference on artificial intelligence},
  volume={36},

  pages={4120--4127},
  year={2022}
}

@inproceedings{li2022tirgn,
  title={TiRGN: Time-Guided Recurrent Graph Network with Local-Global Historical Patterns for Temporal Knowledge Graph Reasoning.},
  author={Li, Yujia and Sun, Shiliang and Zhao, Jing},
  booktitle={IJCAI},
  pages={2152--2158},
  year={2022}
}

@article{zhu2021neural,
  title={Neural bellman-ford networks: A general graph neural network framework for link prediction},
  author={Zhu, Zhaocheng and Zhang, Zuobai and Xhonneux, Louis-Pascal and Tang, Jian},
  journal={Advances in neural information processing systems},
  volume={34},
  pages={29476--29490},
  year={2021}
}

@inproceedings{teru2020inductive,
  title={Inductive relation prediction by subgraph reasoning},
  author={Teru, Komal and Denis, Etienne and Hamilton, Will},
  booktitle={International conference on machine learning},
  pages={9448--9457},
  year={2020},
  organization={PMLR}
}

@article{liu2021indigo,
  title={Indigo: Gnn-based inductive knowledge graph completion using pair-wise encoding},
  author={Liu, Shuwen and Grau, Bernardo and Horrocks, Ian and Kostylev, Egor},
  journal={Advances in Neural Information Processing Systems},
  volume={34},
  pages={2034--2045},
  year={2021}
}

@inproceedings{lee2023ingram,
  title={InGram: Inductive knowledge graph embedding via relation graphs},
  author={Lee, Jaejun and Chung, Chanyoung and Whang, Joyce Jiyoung},
  booktitle={International Conference on Machine Learning},
  pages={18796--18809},
  year={2023},
  organization={PMLR}
}

@inproceedings{lee2023temporal,
  title={Temporal Knowledge Graph Forecasting Without Knowledge Using In-Context Learning},
  author={Lee, Dong-Ho and Ahrabian, Kian and Jin, Woojeong and Morstatter, Fred and Pujara, Jay},
  booktitle={Proceedings of the 2023 Conference on Empirical Methods in Natural Language Processing},
  pages={544--557},
  year={2023}
}

@inproceedings{gastinger2023comparing,
  title={Comparing apples and oranges? on the evaluation of methods for temporal knowledge graph forecasting},
  author={Gastinger, Julia and Sztyler, Timo and Sharma, Lokesh and Schuelke, Anett and Stuckenschmidt, Heiner},
  booktitle={Joint European conference on machine learning and knowledge discovery in databases},
  pages={533--549},
  year={2023},
  organization={Springer}
}

@inproceedings{han2021learning,
  title={Learning neural ordinary equations for forecasting future links on temporal knowledge graphs},
  author={Han, Zhen and Ding, Zifeng and Ma, Yunpu and Gu, Yujia and Tresp, Volker},
  booktitle={Proceedings of the 2021 conference on empirical methods in natural language processing},
  pages={8352--8364},
  year={2021}
}

@inproceedings{liao2024gentkg,
  title={GenTKG: Generative Forecasting on Temporal Knowledge Graph with Large Language Models},
  author={Liao, Ruotong and Jia, Xu and Li, Yangzhe and Ma, Yunpu and Tresp, Volker},
  booktitle={NAACL-HLT (Findings)},
  year={2024}
}

@inproceedings{zhangtrix,
  title={TRIX: A More Expressive Model for Zero-shot Domain Transfer in Knowledge Graphs},
  author={Zhang, Yucheng and Bevilacqua, Beatrice and Galkin, Mikhail and Ribeiro, Bruno},
  booktitle={The Third Learning on Graphs Conference},
  year={2025}
}

@inproceedings{zhu2021learning,
  title={Learning from history: Modeling temporal knowledge graphs with sequential copy-generation networks},
  author={Zhu, Cunchao and Chen, Muhao and Fan, Changjun and Cheng, Guangquan and Zhang, Yan},
  booktitle={Proceedings of the AAAI conference on artificial intelligence},
  volume={35},
  number={5},
  pages={4732--4740},
  year={2021}
}

@inproceedings{xu2020inductive,
  title={Inductive Representation Learning on Temporal Graphs},
  author={Xu, Da and Ruan, Chuanwei and Korpeoglu, Evren and Kumar, Sushant and Achan, Kannan},
  booktitle={International Conference on Learning Representations},
  year={2020}
}

@article{rossi2020temporal,
  title={Temporal Graph Networks for Deep Learning on Dynamic Graphs},
  author={Rossi, Emanuele and Chamberlain, Ben and Frasca, Fabrizio and Eynard, Davide and Monti, Federico and Bronstein, Michael},
  journal={arXiv preprint arXiv:2006.10637},
  year={2020}
}

@misc{pan2026fittervocabularyagnosticcrossdomaininference,
  title={FITTER: Vocabulary-Agnostic Cross-Domain Inference on Temporal Knowledge Graphs},
  author={Jiaxin Pan and Mojtaba Nayyeri and Osama Mohammed and Daniel Hernandez and Rongchuan Zhang and Cheng Cheng and Steffen Staab},
  year={2026},
  eprint={2608.10668},
  archivePrefix={arXiv},
  primaryClass={cs.AI},
  url={https://arxiv.org/abs/2608.10668},
}

@inproceedings{kim2026flock,
  title={Flock: A knowledge graph foundation model via learning on random walks},
  author={Kim, Jinwoo and Huang, Xingyue and Olejniczak, Krzysztof and Min, Kyungbin and Bronstein, Michael and Hong, Seunghoon and Ceylan, Ismail I},
  booktitle={International Conference on Learning Representations},
  volume={2026},
  pages={36664--36705},
  year={2026}
}
